\documentclass{article}
\usepackage{arxiv}

\usepackage[utf8]{inputenc} 
\usepackage[T1]{fontenc}    
\usepackage{hyperref}       
\usepackage{url}            
\usepackage{booktabs}       
\usepackage{amsfonts}       
\usepackage{nicefrac}       
\usepackage{microtype}      
\usepackage[ruled]{algorithm2e}

\usepackage{wrapfig}
\usepackage{enumitem}
\setenumerate{leftmargin=5mm}
\setitemize{leftmargin=5mm}

\usepackage{natbib}
\usepackage{times}
\usepackage{microtype}
\usepackage{graphicx}
\usepackage{amsmath}
\usepackage{subfig}
\usepackage{booktabs} 

\newtheorem{theorem}{Theorem}[section]

\newtheorem{defn}[theorem]{Definition}
\newtheorem{prop}[theorem]{Proposition}
\newtheorem{lemma}{Lemma}[section]

\hypersetup{
	colorlinks,
	linkcolor={red!50!black},
	citecolor={blue!50!black},
	urlcolor={blue!80!black}, 
}
\usepackage{etoolbox}

\renewcommand{\vec}[1]{\boldsymbol{#1}} 
\newcommand{\mat}[1]{\boldsymbol{#1}} 

\newcommand{\inv}[0]{^{-1}} 

\newcommand{\prob}{\mathrm{Pr}}
\newcommand{\flow}{\mathcal{F}}
\newcommand{\copulaflow}{\mathcal{C}}

\newcommand{\range}[1]{\mathbf{ran}({#1})}
\newcommand{\domain}[1]{\mathbf{dom}({#1})}
\newcommand{\uniform}[2][0]{\text{Uniform}\left[#1, #2\right]}
\usepackage{amsfonts}
\usepackage{xcolor}

\usepackage[colorinlistoftodos,prependcaption,textsize=tiny]{todonotes}

\title{Copula Flows for Synthetic Data Generation}

\author{
 Sanket Kamthe \\
 Department of Computing \\
  Imperial College London\\
  JP Morgan Chase CTO-Applied Research\\
  
  \texttt{s.kamthe15@imperial.ac.uk} \\
   \And
 Samuel Assefa \\
  JP Morgan Chase AI Research\\

  \texttt{samuel.a.assefa@jpmorgan.com } \\
  \And
 Marc Deisenroth\\
  Department of Computer Science\\
  University College London\\
 
  \texttt{m.deisenroth@ucl.ac.uk} \\
  }

\begin{document}

\maketitle

\begin{abstract}
The ability to generate high-fidelity synthetic data is crucial when available
(real) data is limited or where privacy and data protection standards allow only
for limited use of the given data, e.g., in medical and financial data-sets.  
%
Current state-of-the-art methods for synthetic data generation are based on
generative models, such as Generative Adversarial Networks (GANs). Even though
GANs have achieved remarkable results in synthetic data generation, they are
often challenging to interpret. Furthermore, GAN-based methods can suffer when
used with mixed real and categorical variables. Moreover, loss function
(discriminator loss) design itself is problem specific, i.e., the generative
model may not be useful for tasks it was not explicitly trained for.  
In this paper, we propose to use a probabilistic model as a synthetic data
generator. Learning the probabilistic model for the data is equivalent to
estimating the density of the data.  Based on the copula theory, we divide the
density estimation task into two parts, i.e., estimating univariate marginals
and estimating the multivariate copula density over the univariate marginals. We use
normalising flows to learn both the copula density and univariate marginals.  
We benchmark our method on both simulated and real data-sets in terms of density
estimation as well as the ability to generate high-fidelity
synthetic data.

\end{abstract}

\section{Introduction}
\label{intro}

Machine learning is an integral part of daily decision-support systems, ranging from
personalised medicine to credit lending in banks and financial institutions. The
downside of such a pervasive use of machine learning is the privacy concern
associated with the use of personal data ~\citep{Elliot2018}. In some
applications, such as medicine, sharing data with privacy in mind is fundamental
for advancing the science~\citep{Beaulieu2019}. However, the available data may
not be sufficient to realistically train state-of-the-art machine learning
algorithms. Synthetic data is one way of mitigating this challenge. 



Current state-of-the-art methods for synthetic data generation, such as
Generative Adversarial Networks (GANs)~\citep{Goodfellow2014}, use complex deep
generative networks to produce high-quality synthetic data for a large variety
of problems~\citep{Choi2017, Xu2019}. 
However, GANs are known to be unstable and delicate to
train~\citep{Arjovsky2017}. Moreover, GANs are also notoriously difficult to
interpret, and finding an appropriate loss function is an active area of
research~\citep{wang2019}.  Another class of models, variational auto encoders
(VAEs)~\citep{Kingma2013, Rezende2015} have also been used to generate data,
although their primary focus is finding a lower-dimensional data representation.
While deep generative models, such as  VAEs and GANs,  can generate
high-fidelity synthetic data, they are both difficult to interpret. Due to the
complex composition of deep networks, it is nearly impossible to characterise
the impact of varying weights on the density of the generated data. For VAEs, a
latent space may be used to interpret different aspects of the data. 

In this work, we focus on probabilistic models based on copulas. Sklar showed
that any multivariate distribution function can be expressed as a function of
its univariate marginals and a copula~\citep{Sklar1959}. 
However, estimating copulas for multivariate mixed data types is typically hard,
owing to 1) very few parametric copulas have multivariate formulations and 2)
copulas are not unique for discrete or count data~\citep{Nikoloulopoulos2009}.
To address the first challenge, the current standard is to use a pairwise copula
construction~\cite{Aas2009}, which involves two steps identifying the pair copula
and a tree structure defining pairwise relationship~\citep{Elidan2010,
Lopez-paz2013, Czado2010, Chang2019}. 

\begin{figure*}[tb]
\centering
\scalebox{0.8}{
	\begin{tikzpicture}
	\centering
		\node[inner sep=0pt] (copula) at (0,0)
		{\includegraphics[width=.30\textwidth]{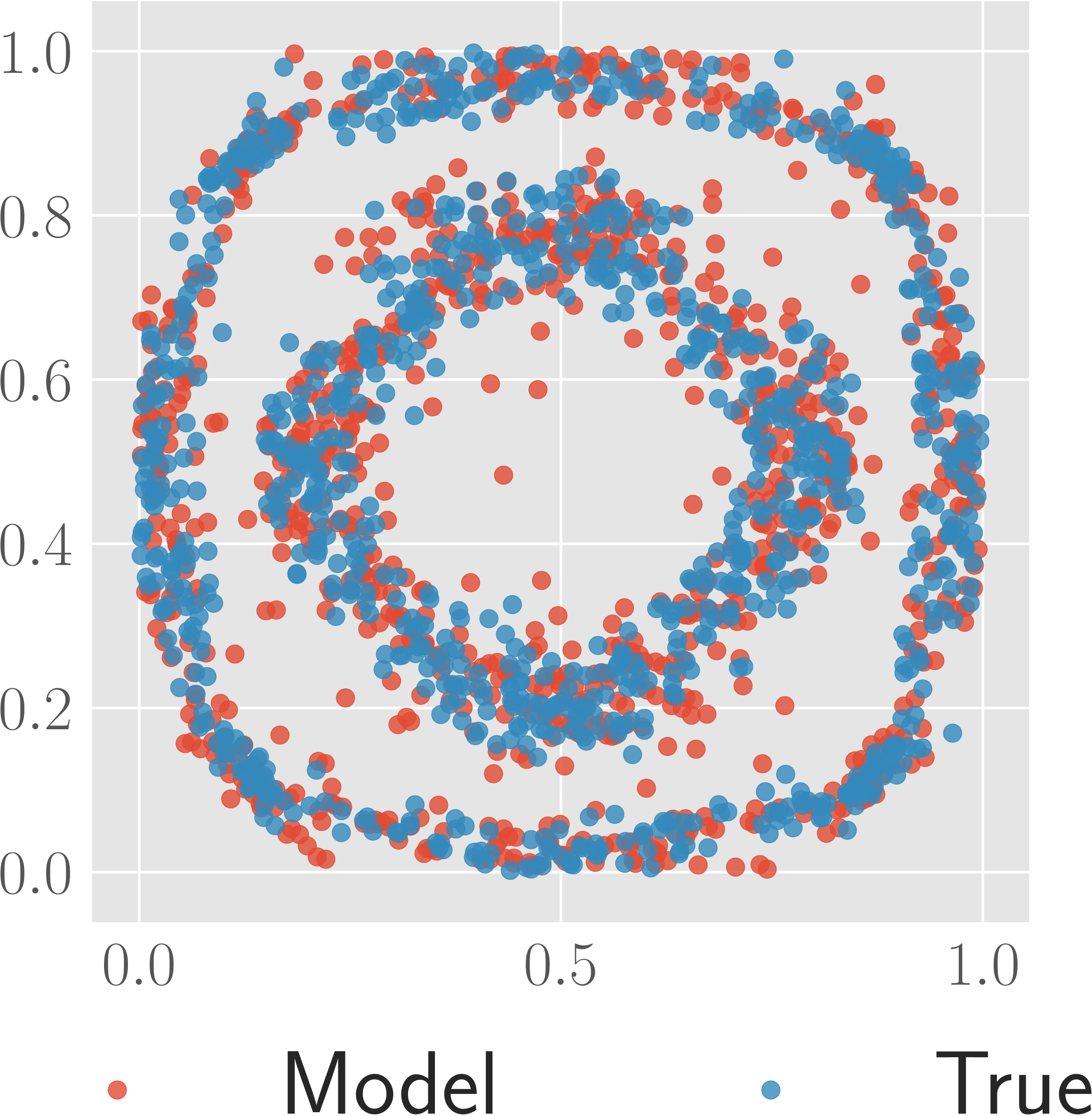}};
		\node[inner sep=0pt] (plus) at (2.5,0)
		{\includegraphics[width=.02\textwidth]{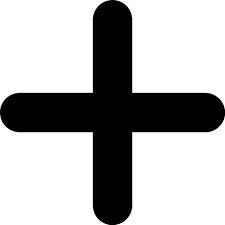}};
		\node[inner sep=0pt] (marginal) at (5,0)
		{\includegraphics[width=.30\textwidth]{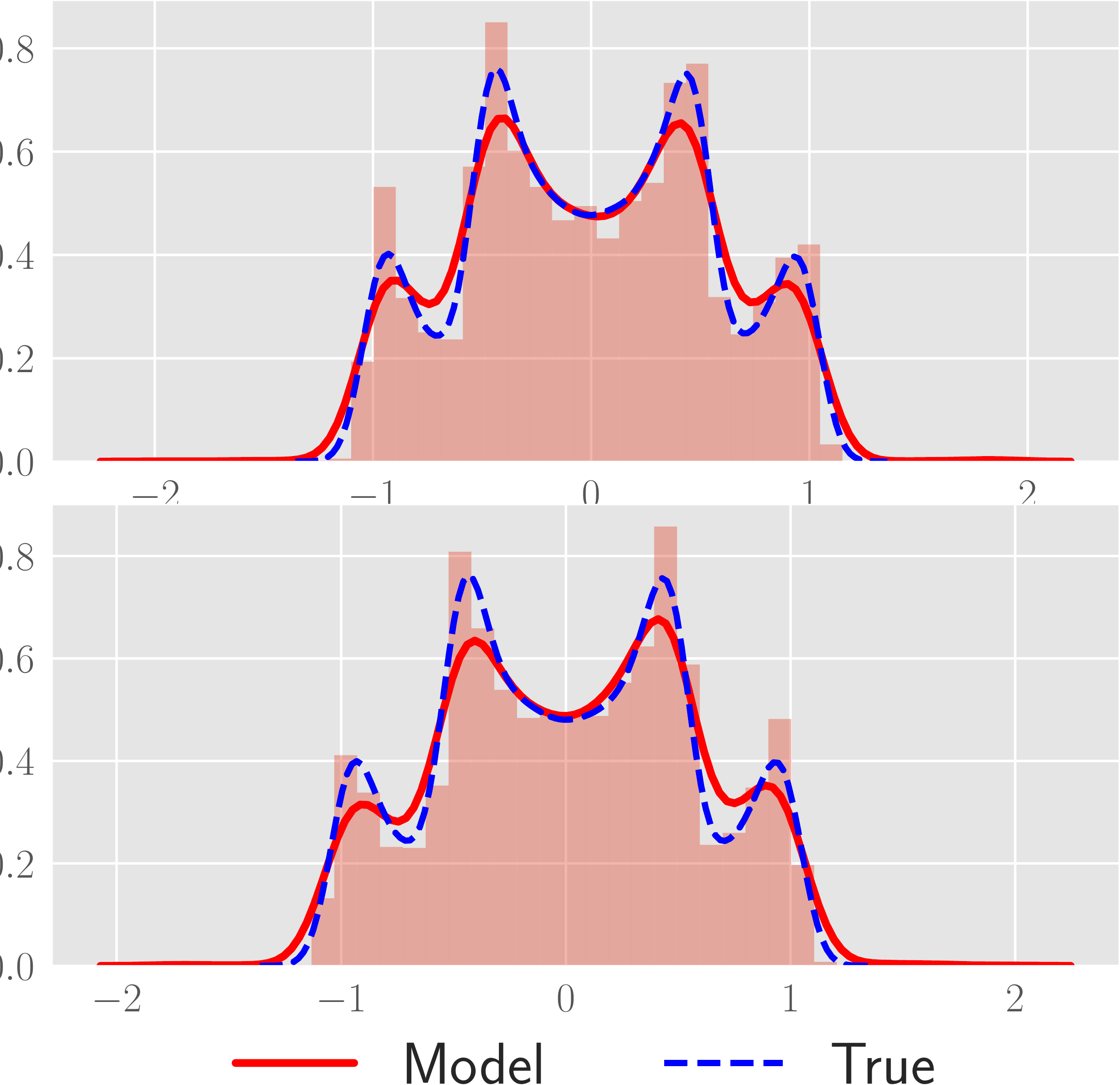}};
		\node[inner sep=0pt] (equal) at (7.6,0)
		{\includegraphics[width=.02\textwidth]{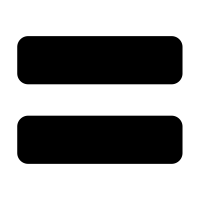}};
		\node[inner sep=0pt] (joint) at (10,0.0)
		{\includegraphics[width=.30\textwidth]{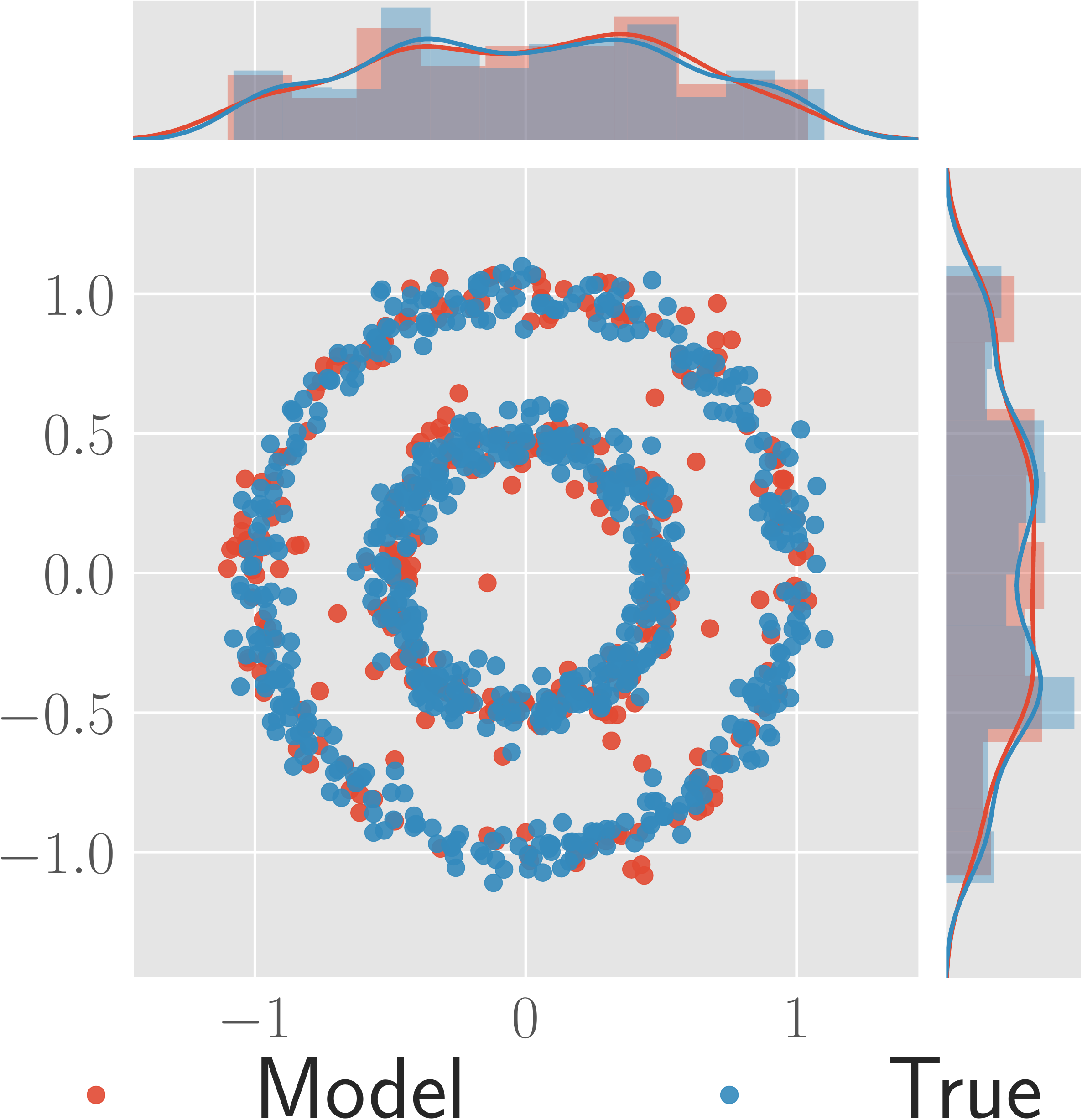}};
		
		\node[inner sep=0pt] (coupalHead) at (0,-2.5) {a) Copula flow };
		
		\node[inner sep=0pt] (MargHead) at (5,-2.5) {b) Marginal flow};
		
		\node[inner sep=0pt] (MargHead) at (10,-2.5) {b) Complex joint density
		};
	\end{tikzpicture}
	}
	\caption{Learning the joint and marginal distributions for `2 rings' data.
	 This figure demonstrates how copulas
	 and marginals can be combined to obtain complex joint distributions.
	The copula associated with the `2 rings' distribution cannot be modelled
	by any standard bi-variate copula, whereas our proposed copula flow method
	can learn this structure.}
	\label{fig:copulaintro}
\end{figure*}
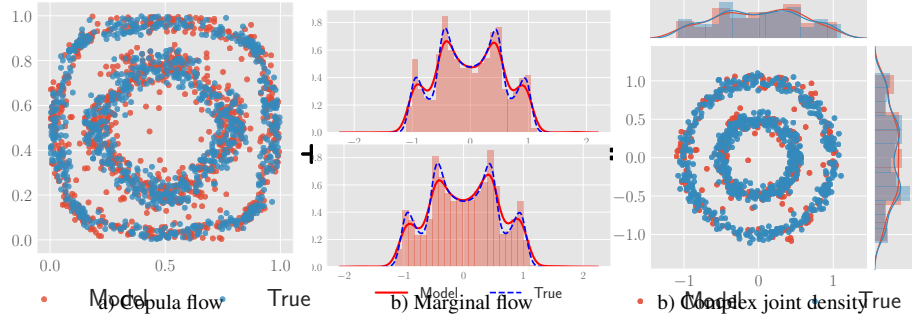

In this
paper, we propose a probabilistic synthetic data generator that is interpretable
and can model arbitrarily complex data-sets. 
Our approach is based on normalising flows to estimate
copulas. We show that our model is flexible enough to learn complex
relationships and does not rely on explicit learning of the graph or parametric
bivariate copulas.  Figure~\ref{fig:copulaintro} illustrates the
flexibility of our proposed method by modelling copula with a complex structure.
We are not aware of any copula based method able to learn the copula shown in
Figure~\ref{fig:copulaintro} a).  

To deal with count data, we propose a modified spline flow~\citep{Durkan2019} as
a continuous relaxation of discrete probability mass functions. This formulation
allows us to build complex multivariate models with mixed data types that can
learn copula-based flow models, which can be used for the following tasks:
\begin{itemize}
	\item [1.] \textit{Synthetic data generation:} Use the estimated model to
	generate new data-sets that have a distribution similar to the training set.
	
	\item[2.] \textit{Inferential statistics:} When the copula has
	learned the relationship between the variables correctly, one can change the
	marginals to study the effects of change. For example, if we estimated a
	copula flow based on a UK dataset with age as one of the variables, we can
	modify the marginal of the age distribution to generate synthetic data for a
	different country. To generate data for Germany, we would use
	Germany's age distribution as a marginal for the data-generating process.
	\item[3.]\textit{Privacy preservation:} The data generated from the copula
	flow model is fully synthetic, i.e., the generator does not rely on actual
	observations for generating new data. One can perturb the generated data
	based on differential privacy mechanisms to prevent privacy attacks based on
	a large number of identical queries.
\end{itemize}

\section{Background}
\label{sec:background}
We denote the domain of a function $ F $ by $ \domain F $ and range by $\range F$. 
We use capital letters $ X, Y $ to denote random variables 
and lower case $ x, y $ to represent their realisations. Bold symbols $ \mat X =
[X_1, \ldots, X_d]$ and  corresponding $ \vec x = [x_1, \ldots, x_d]$ represent vectors of random  variables and their realisations, respectively.  
We denote the distribution function (also known as the Cumulative Distribution
Function (CDF)) of a random variable $X$ by $F_X$, and the corresponding Probability Density Function (PDF) by $f_X$.
Such functions play an important role in copula theory. If we map a
random variable through its own distribution function we get an uniformly distributed random variable-$
\uniform{1}$ or uniform marginal. This is known as
\emph{probability integral transform}~\citep{David1948}.\footnote{This
definition holds only for continuous variables; we discuss CDFs for discrete variables in section \ref{sec:discreteFlows}} A copula
is a relationship, literally a link, between uniform marginals obtained via the
probability integral transform~\citep{Nelsen2006}. Two random variables are
independent if the joint copula of their uniform marginals is
independent. Conversely, random variables are \textit{linked} via their
copulas. 


\subsection{Copulas}
 For a pair of random variables, the marginal CDFs $F_X, F_Y$ describe how each
variable is individually distributed. The joint CDF $F_{X,Y}$ tells us how they
are jointly distributed. Sklar's seminal result \citep{Sklar1959} allows us to
write the joint CDF as a function of the univariate marginals $ F_X $ and $ F_Y
$, i.e., 
		\begin{align}
			F_{X, Y}(X, Y) &= C \left( U_X, U_Y \right) = C \left(F_X (X), F_Y(Y) \right),
		\end{align}
		where $C$ is a copula. 
Here, the
uniform marginals are obtained by the probability integral transform, i.e., $ U_X =
F_X(X)$ and $U_Y = F_Y(Y)$.

Copulas describe the dependence structure independent of the marginal
distribution, which we exploit for synthetic data generation; especially in
privacy preservation, where we perturb data samples by another random process,
e.g., the Laplacian mechanism~\citep{Dwork2013}. This changes the marginal
distribution but crucially it does not alter the copula of the original data.

For continuous marginals $ F_X $ and $ F_Y $, the copula $C$ is unique, whereas
for discrete marginals it is unique only on $ \range{F_X} \times \range{F_Y}$.
A corollary of Sklar's theorem is that $X$ and $Y$ are independent if
and only if their copula is an independent copula, i.e., $C (F_X(X), F_Y(Y))= F_X(X) \, F_Y(Y)$. 
\subsection{Generative Sampling}
\label{sec:gen_sampling}
We can use the constructive inverse of Sklar's theorem for building a generative model.
The key concept is that the inverse of the probability integral transform,
called \textit{inverse transform sampling}, allows us to generate random samples
starting from a uniform distribution.  The procedure to generate a random sample $
x $ distributed as $ F_X $ is to first sample a random variable $ u \sim
\uniform{1}$ and second to set $ x :=  F_X^{(-1)} (u)$.
 Here, the the function $ F^{(-1)}  $ is a quasi-inverse function.
 \begin{defn}[Quasi-inverse function] 
	\label{defn:quasi-inverse}
	Let $ F $ be a CDF. Then
 	the function $ F^{(-1)} $ is a quasi-inverse function of $F$ with domain $
 	[0, 1] $ such that 
 	 $F(F^{(-1)} (y)) = y \quad \forall \, y \in \range F $
 	and $ F^{(-1)} (y)  = \inf \left\lbrace x | F(x) \geq y \right\rbrace
 		\forall \, x \in \domain F, \, \text{and} \, y \notin \range F$.
	For strictly monotonically increasing $ F $, the quasi-inverse becomes the regular inverse CDF $
 	F^{-1}$~\citep{Nelsen2006}.
 \end{defn}

The inverse function, also called as the \emph{quantile function}, maps a random
variable from a uniform distribution to $F$-distributed random variables, i.e.,
$ X = F\inv (U) $ or $ F\inv : \uniform{1} \rightarrow \range F $.  Hence, we
can present the problem of synthetic data generation as the problem of
estimating the (quasi)-inverse function for the given data. 

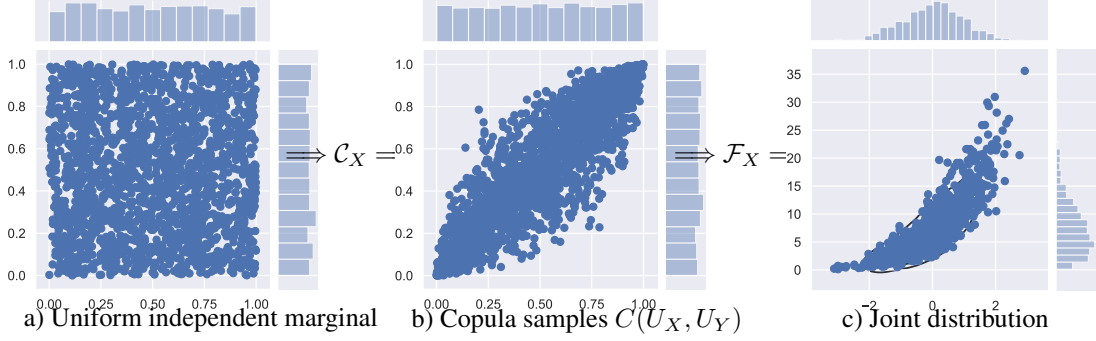
\begin{figure*}
	\begin{tikzpicture}
	\centering
		\node[inner sep=0pt] (uniform) at (-0.5,0)
		{\includegraphics[width=.25\textwidth]{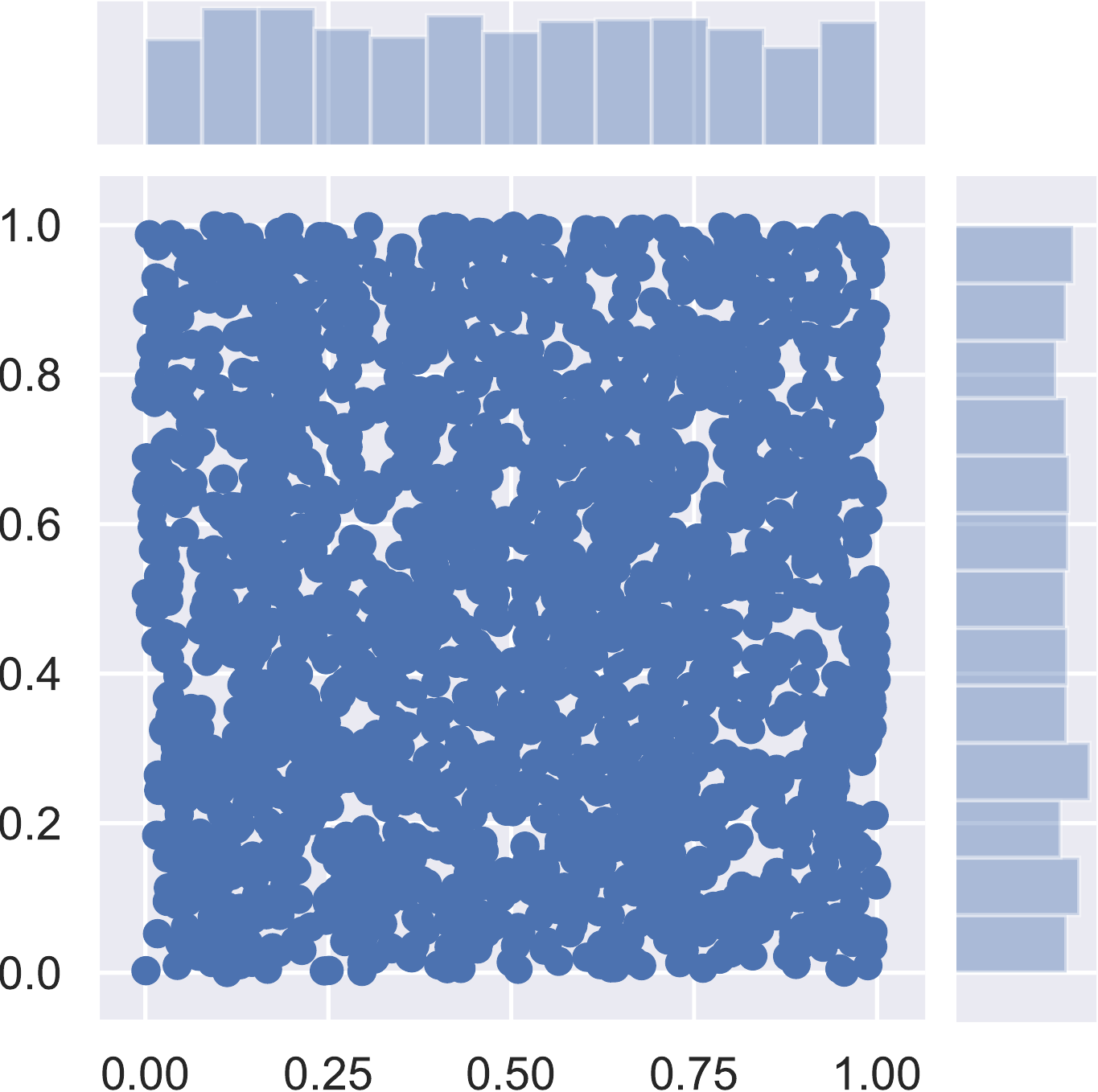}};
		\node[inner sep=0pt] (copula_text) at (2.0,0)
		{$\Longrightarrow $ $\copulaflow_X$ $\Longrightarrow $};
		\node[inner sep=0pt] (marginal) at (4.65, 0)
		{\includegraphics[width=.25\textwidth]{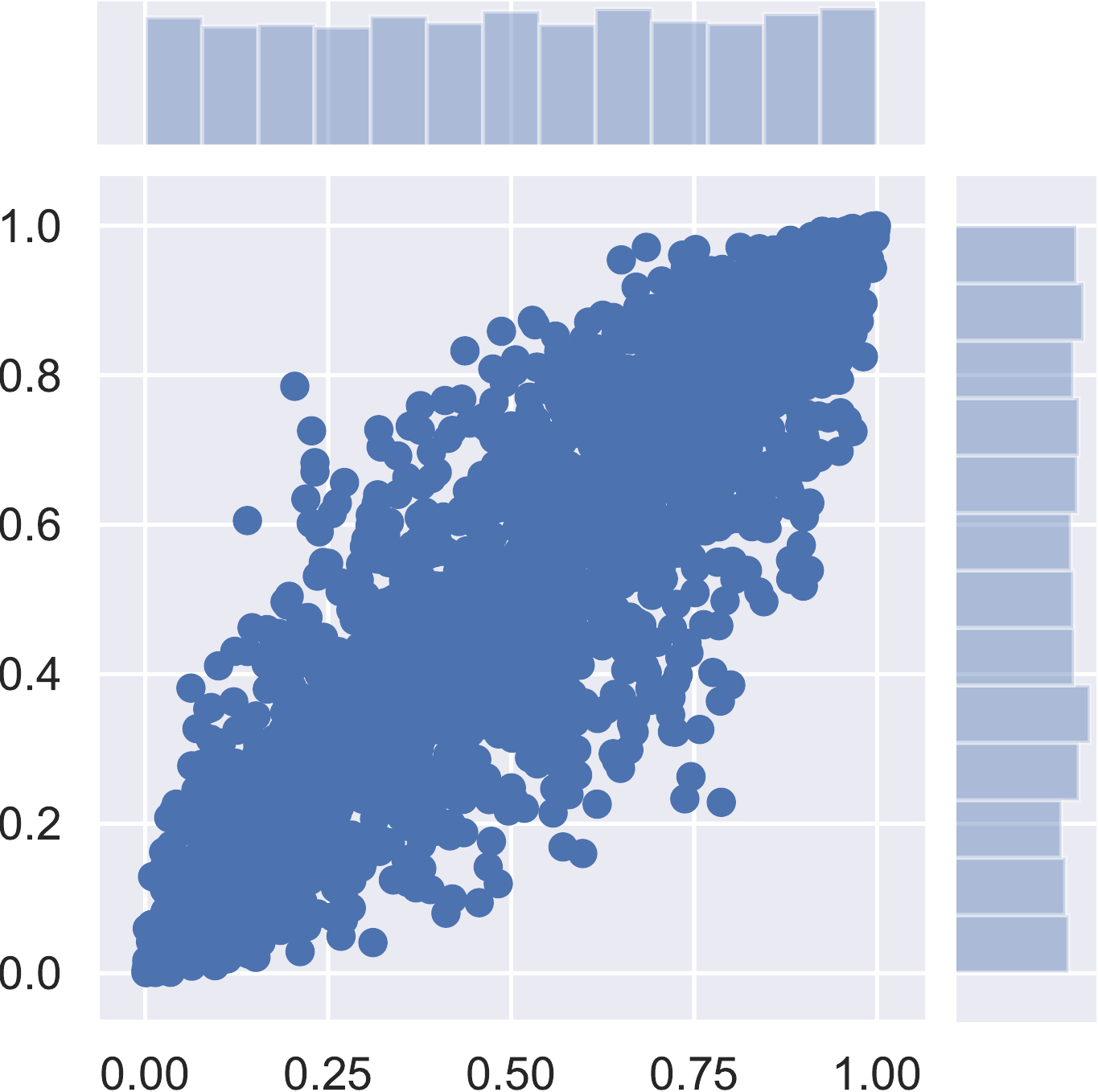}};
		\node[inner sep=0pt] (equal) at (7.2,0)
		{$\Longrightarrow $ $\flow_X$ $\Longrightarrow $};
		\node[inner sep=0pt] (joint) at (9.85,0)
		{\includegraphics[width=.25\textwidth]{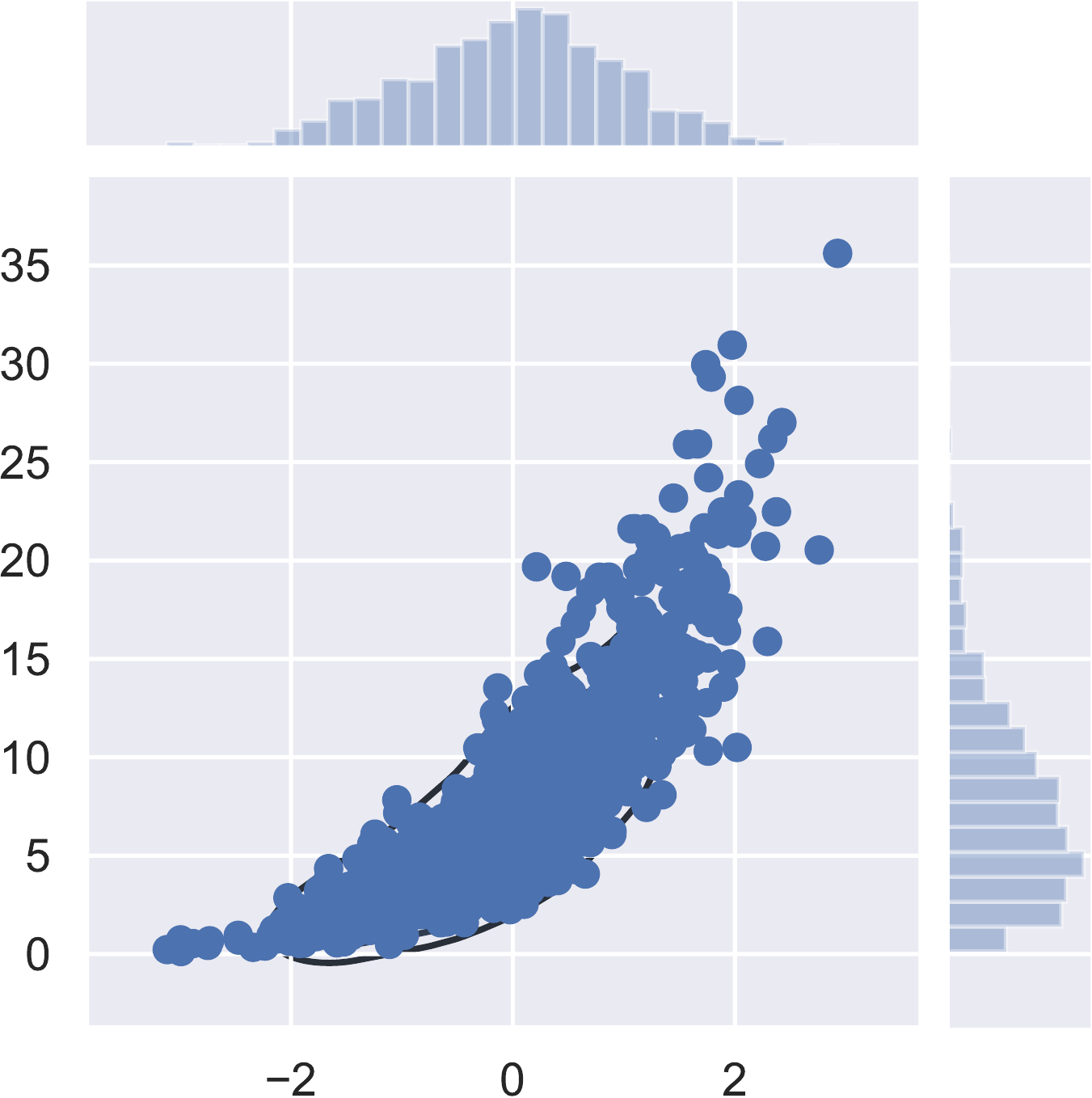}};
		
		\node[inner sep=0pt] (UnilHead) at (0,-2.2) {a) Uniform independent
		marginal};
		
		\node[inner sep=0pt] (copulaHead) at (5,-2.2) {b) Copula samples $C (U_X, U_Y)$ };
		
		\node[inner sep=0pt] (MargHead) at (9.9,-2.2) {c) Joint distribution
		};

	\end{tikzpicture}
	\caption{Copula flow generative model. Left: We start with uniformly distributed
	 independent variable samples.  These samples are passed through the copula
	 flow network to generate correlated uniformly distributed random variables (middle).
	 The correlated variables are transformed to the desired distribution by
	 using univariate marginal flows $\flow_X$ (right).}
	 \label{fig:generative_model}
\end{figure*}

To generate samples from $F_X$ using a copula-based model, we start with a pair of uniformly distributed random
variables, then pass them through inverse of the copula CDF to obtain correlated
uniform marginals; see Figure~\ref{fig:generative_model} b). Note that the
copulas are defined over uniform marginals, i.e., the random variables are
uniformly distributed. We use the correlated univariate marginals to subsequently generate
$F$-distributed random variables via inverse transform sampling; see
Figure~\ref{fig:generative_model} c). Hence, if we can learn the CDFs for the
copula as well as the marginals, we can then combine these two models to
generate correlated random samples that are distributed similar to the training
data. The procedure is illustrated in
Figure~\ref{fig:generative_model}.

\section{Copula Flows}

The inverse function used to generate synthetic data can be described as a flow
function $ \flow_X \approx F_X^{(-1)}$ that transforms a uniform random variable
$ U$ into $ X\sim F_X$, see Figure~\ref{fig:generative_model}. We can interpret the (quasi)-inverse function as a
normalising flow~\citep{Tabak2010, Rezende2015, Papamakarios2019} that
transforms a uniform density into the one described by the copula distribution
function $C$. We refer it as \emph{copula flow}. 

\subsection{Normalising Flows}
Normalising flows are compositions of smooth, invertible mappings that
transform one density into another~\citep{Rezende2015,Papamakarios2019}. The
approximation in $\flow_X \approx F_X^{(-1)}$ indicates that we are estimating a
(quasi) inverse distribution function by using a flow function $\flow_X  $. If we
learn the true CDF $ F_X$, we
have
\begin{align}
	F_X(x) = \flow_X\inv (x) = u, \, u\sim \uniform{1},   \quad \forall x \in \domain {F_X},
\end{align} 
due to uniqueness of the CDF.
An advantage of the normalising flow formulation is that we can construct
arbitrarily complex maps as long as they are invertible and
differentiable.  

Consider an invertible flow function $ \vec\flow :  \mathbb{R}^{D}
\rightarrow \mathbb{R}^{D}$, which transforms a random variable as $ \vec X=
\vec \flow (\vec U) $. By using the change of
variables formula, the density $f_{\vec X}$ of variable $\vec X$ is obtained as
\begin{align}
\label{eq:normFlow}
f_{\vec X}(\vec X) &= f_{\vec U}\left(\flow\inv (\vec X)\right) \left|\operatorname{det} \left(\tfrac{\partial  \vec \flow\inv(\vec X)}{\partial \vec X} \right) \right| 
 =  \,  \left|\operatorname{det} \left(\tfrac{\partial  \vec \flow\inv(\vec X)}{\partial \vec X} \right) \right|,
\end{align}
where $f_{\vec U}\left(\flow\inv (\vec X)\right) = f_{\vec U}(\vec U) = 1.0$, since $f_{\vec U}=\uniform{1}$.


For the copula flow model, we assume that the copula density $c$ of the copula CDF $C$ exists. Starting with the bivariate case for random variables $X, Y$, we compute the density $f_{XY}$
via the partial derivatives of the $C$, i.e., 
\begin{align}
\label{eq:bivariate_density}
	f_{XY} &= \frac{\partial^2 C (F_X, F_Y) }{\partial F_X \partial F_Y} 
	= c_{XY}(F_X, F_Y)  \, f_X \, f_Y.
\end{align}
We can generalise this result to the joint density $f_{\mat
X}(\vec X) $ of a $d$-dimensional random  vector  $\vec X =[ X_{1}, \ldots,
X_{d}]$ as
\begin{align}
\label{eqn:jointMarginal}
f_{\mat X}&= 
\underbrace{
c_{\mat X} (F_{X_1}, \ldots, F_{X_d}) 
\vphantom{\prod\nolimits_{k=1}^{d}}
}_{\text{copula density}} 
\, 
\underbrace{
\prod\nolimits_{k=1}^{d}f_{X_k}
}_{\text{marginal density}}
.
\end{align}

To construct the copula-based flow model, we begin with rewriting the joint
density $f_{\mat X}$ in~\eqref{eqn:jointMarginal} in terms of marginal
flows $ \flow $ and the joint copula flow $\copulaflow_{\vec X}$. We then obtain
\begin{align}
	\label{eqn:jointMarginalNoText}
	f_{\mat X}(\vec X)
	 &=\left|\operatorname{det} \left(\tfrac{\partial  \copulaflow_{\vec X}\inv(\vec U_{\vec X})}
	{\partial \vec U_{\vec X}} \right) \right| \prod\nolimits_{k=1}^{d} 
	\left| \tfrac{\partial  \flow_{X_k}\inv(X_k)}{\partial X_k}  \right|,
\end{align}
where $\copulaflow_{\vec X}$ is the copula flow function and $\flow_{X_1},
 \ldots, \flow_{X_d}$ are marginal flow functions; see Appendix~\ref{sec:apendixFlow}
 for the detailed derivation. Even though the copula itself is defined on
 independent univariate marginals functions, we use $X$ in the notation to
 emphasise that these univariates are obtained by inverting the flow $U_{X_k} =
 \flow\inv _{X_k}(X_k)$ for marginals.

Given a data-set $\{\vec x_1, \ldots, \vec x_N\}$ of size $N$, we 
train the flow by maximising the total log-likelihood $ 
\mathcal L=\log f_{\mat X}(\vec x) = \sum_{n=1}^{N} \log
f_{\mat X}(\vec x_{n})$ with respect to the parametrisation of flow function
$ \flow $.  
The log-likelihood can be written as the sum of two terms, namely
\begin{align}
\label{eqn:Loglikelihood}
\mathcal{L} 
&=\mathcal{L}_{\copulaflow_{\vec X}} 
+\mathcal{L}_{\vec \flow} 
= 
\mathcal{L}_{\copulaflow_{\vec X}} 
+ \sum\nolimits_{k=1}^d \mathcal{L}_{X_k}.
\end{align}
The copula flow log-likelihood $
 \mathcal{L}_{\copulaflow_{\vec X}} $ depends on the marginals flows. Hence, we
 train the marginal flows first and then the copula flow. The gradients of
 both log-likelihood terms are independent as they are separable. The procedure
 of first training the marginals before fitting copula models is standard in the
 copula literature and yields better performance and numerical stability
 \citep{Joe2005}. 


\subsection{Copula Flows}
Copula is a CDF defined over uniform marginals, i.e., $\uniform{1}^d \rightarrow
\uniform{1}$. We can use the inverse of this CDF to generate samples via inverse
transform sampling. For the multivariate case, we can use the conditional
generation procedure~\citep{Hansen1994}. Let $ \vec U_{\vec X} = [U_{X_1}, \ldots,
U_{X_d}] $ be a random vector with copula $ \mat C_{\vec X} $, and $   \vec U=
U_1, \ldots, U_d$ be i.i.d.  $ \uniform{1} $  random variables. Then the
multivariate flow transform $\vec  U_X :=  \mat \copulaflow_{\vec X}  ( \vec U)
$ can be defined recursively as
\begin{align}
\label{eqn:multivariateCopula}
	U_{X_1} &:= \copulaflow_{X_1} (U_{1}),\quad
	U_{X_k} := \copulaflow_{X_{k|1, \ldots k-1}}\left( U_{k} | U_{X_1, \ldots, X_{k-1}} \right),
	  \quad 2 \leq k \leq d,
\end{align}
where $ \copulaflow_{X_{k|1, \ldots k-1}} = C\inv_{X_{k|1, \ldots k-1}} $ is the
flow function conditioned on all the variables $ U_{X_1}, \ldots,
U_{X_{k-1}}$. The key concept is the interpretation of the inverse of the
(normalising) copula flow $\copulaflow_{\vec X}\inv$ as a conditional CDF.
Moreover, we estimate $ \copulaflow_{X_{k|1, \ldots k-1}}$ recursively, with one
dimension at a time,  via a neural spline
network~\citep{Durkan2019}. The conditioning variables, $X_{1}, \ldots X_{k-1}$ are
input to the network that outputs spline parameters for the flow $
\copulaflow_{X_{k|1, \ldots k-1}}$.
The procedure above is similar to
auto-regressive density estimators~\citep{Papamakarios2017,Papamakarios2019}
with Masked Autoregressive Flow (MAF)
structure~\citep{Germain2015,Papamakarios2017}. Similar to MAF we can stack such
multiple flows to create flexible multivariate copula flow $\mat
\copulaflow_{\vec X}$. In the copula literature this multivariate extension is
convenient for Archimedean copulas~\citep{Nelsen2006}, where the generators for
such copulas can be interpreted as  conditional flow functions. 


\subsection{Univariate Marginal Flows}
To estimate univariate marginals, we can use both parametric and non-parametric
density estimation methods as the dimensionality is not a challenge. However,
for training the copula, we require models that can be inverted easily to obtain
uniform marginals, i.e., we need to have a well-defined CDF for the methods we
employ for the density estimation. Simultaneously, we need a model that can be used
for generating data via inverse transform sampling.
We propose to use monotone rational quadratic splines, neural spline flows
(NSF)~\citep{Durkan2019}. With the a sufficiently dense spline, i.e., a large
number of knot positions, we can learn arbitrarily complex marginal
distributions.  As we directly attempt to learn the quasi-inverse
function $ \flow : \uniform{1} \, \rightarrow \, \range F$, where $F$ is the
\textit{true}  CDF, a single parameter vector $ \vec
\theta_d = [\vec \theta_d^w, \vec
\theta_d^h, \vec \theta_d^s]$ describing the width, height and slope parameters,
respectively, is sufficient. 
We document the details of the proposed spline network
in Appendix~\ref{sec:SoftwareDetails}


%


\section{Discrete Data Modelling}
\begin{wrapfigure}[13]{r}{0.35\textwidth}
	\vspace{-15mm}
	   \centering
		\includegraphics[width=0.35\textwidth]{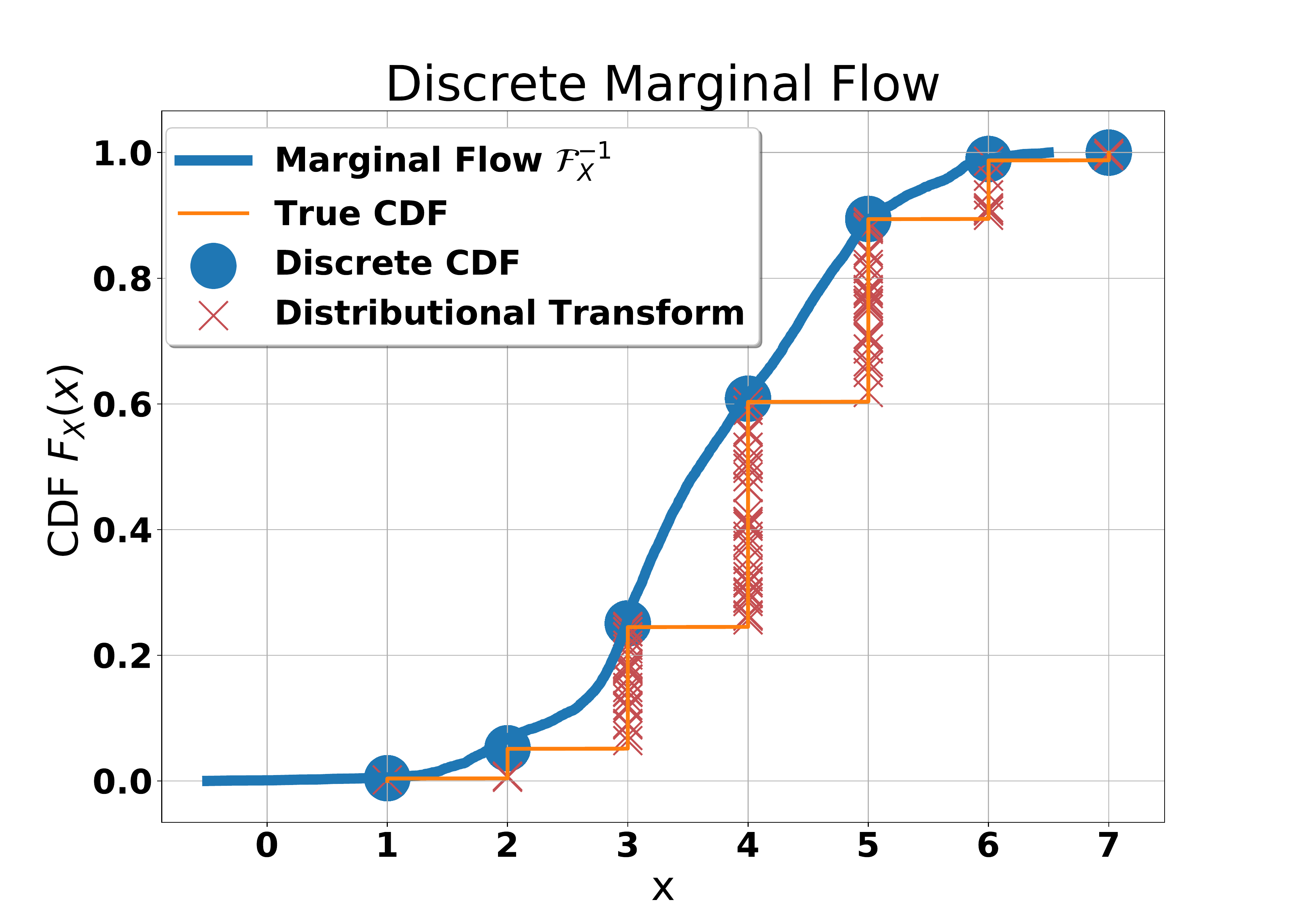}
		\caption{Illustration of discrete marginal flow for three discrete classes. The marginal flow (blue) gives discrete samples when rounded up. The distributional transform
		is uniform along the vertical line of the true CDF (orange). }
		\label{fig:hypergeom}
\end{wrapfigure}
Modelling mixed variables via normalising flows is challenging~\citep{Onken2016,
Ziegler2019, Tran2019}. Discrete data poses challenges for both copula learning
as well as learning flow functions. For marginals, i.e., univariate flow
functions, the input is a uniform distribution continuous in $[0, 1]$, whereas
the output is discrete and hence discontinuous. For copula learning, we need
uniform marginals for the given training data. With discrete inputs to the
inverse flow function (CDF), the output is discontinuous in $[0, 1]$. 

\subsection{Marginal Flows for Discrete/Count Data}
\label{sec:discreteFlows}

We first focus on learning the univariate flow maps, i.e., marginal learning.
Ordinal variables have a natural order, which we can use directly. For
categorical data, we propose to assign each class a unique integer in $\{0
,\ldots, n-1 \}$, where $n  $ is the number of classes. We define a CDF over
these integer values. As this assignment is not
unique, the same category assignment needs to be maintained for training and
data generation.

For discrete data generation, we round the data output of the flow function to
the next higher integer, i.e., we can consider random variable $Y = \text{ceil}
(X)$. However, this procedure does not yield a valid density. To ensure that
that the samples are properly discretised, we use a quantised
distribution~\citep{Salimans2019, Dillon2017} so that density learning can be
formulated as a quantile learning procedure. 

We assign a quantile range for a given class or ordinal integer. We use the same
spline-based flow functions as the one used for continuous marginals, but with
quantisation as the last step. The advantage of this discrete flow is that our
quasi-inverse, i.e., the flow function is a continuous and monotonic function,
which can be trained by maximising the likelihood. Figure~\ref{fig:hypergeom}
shows the spline-based flow function learned for a hypergeometric distribution.
The continuous flow function (blue curve in Figure~\ref{fig:hypergeom}) learned
for the discrete marginal, light orange step function in
Figure~\ref{fig:hypergeom}, allows us to generate discrete data via inverse
transform sampling.

\subsection{Mixed Variable Copulas}

However, the inverse of this flow function,  i.e., the CDF of the marginal,
results in discontinuous values at the locations of the training inputs (blue
circles in Figure~\ref{fig:hypergeom}). Copulas are unique and well defined only
for continuous univariate marginals. An elegant way to find continuous
univariate marginals for the discrete variable is via the distributional
transform~\citep{Ruschendorf2009}.
\begin{defn}[Distributional Transform]
	\label{defn:dist_xform}
	Let $ X \sim F_X $. The modified CDF is defined as 
	\begin{align}
		F_X (x, \lambda) := \prob(X < x) + \lambda \prob(X = x).
	\end{align}
	Let $ V $ be uniformly distributed independent of $X$.  Then the
	distributional transform $ X \rightarrow U $ of $ X $  is 
	\begin{align}
		u := \flow\inv_{X} (x) = F_{X-}(x) + V(F_X(x)- F_{X-}(x)),
	\end{align}
	where $F_{X-}(x) = \prob (X < x)$, $F_{X}(x) = \prob (X \le x)  $ and
	$\prob(X=x) = F_X(x)- F_{-}(x)$. With this, we have $ U  \, \sim \uniform{1}$
	and $ X = \flow_{X} (U)  \, a.s. $~\citep{Ruschendorf2009}. 
\end{defn}
This distributional transform behaves similar to the probability integral
transform for continuous distributions, i.e., a discrete random variable $X$
mapped through the distributional transform gives an uniform marginal. However,
unlike for continuous variables, the distributional map is stochastic and hence not
unique. With the distributional transform, the copula model does not need a
special treatment for discrete variables; see section 2
in~\citep{Ruschendorf2009} for details. In Figure~\ref{fig:hypergeom},
the red cross marks show the values from distributional transform. All  samples
along the $y$-axis that share same $x$-value.
With the distributional transform and marginal splines, the copula flow model
leverages recent advances in normalising flows to learn arbitrarily complex
distributions. Moreover, we can show that:
\begin{theorem}
	A copula flow is a universal density approximator. \quad Proof:
	Appendix~\ref{sec:uni_density_approx}.
\end{theorem}
Hence, we can learn a model to generate any type of data, discrete or
continuous, with the proposed copula flow model. This property holds true when
the flow network converges to the inverse function.

 \section{Related Work}

 The Synthetic Data Vault (SDV) project~\citep{Patki2016} is very close in
 spirit of this paper. In the SDV, the univariate marginals are learned by using
 a Gaussian mixture model, and the multivariate copula is learned as a Gaussian
 copula. 
 Unlike flow-based estimation, proposed in this
 work, copula-based models typically follow a two-step procedure, the
 pair-copula construction or \emph{vine copulas}, by first constructing a
 pairwise dependence structure and then learning pair-copulas for
 it~\citep{Aas2009}. Copula Bayesian networks \citep{Elidan2010} use conditional
 copulas and learn a Bayesian network over the variables. In~\citep{Chang2019},
 the authors propose to use Monte-Carlo tree search for finding the network
 structure. Such models have limited flexibility, e.g., no standard copula model
 can learn the structure in figure~\ref{fig:copulaintro}, which our model
 (copula flow) can learn.
 
 
 GAN-based methods are the workhorse of the majority of recent advances in the
 synthetic data generation, e.g., the conditional GAN~\citep{Xu2019}, builds a
 conditional generator to tackle mixed-variable types, whereas the
 tableGAN~\citep{Park2018} builds GANs for privacy preservation. Unlike the
 tableGAN, the PATE-GAN~\citep{Jordon2019} produces synthetic data with
 differential privacy guarantees. Data generators for medical
 records~\citep{Che2017, Choi2017} are also based on GANs. 
 
 Normalising flow methods have had very rapid growth in terms of the continuous
 improvements on benchmarks introduced by~\citep{Papamakarios2017}. Our work
 relies heavily on the NSF~\citep{Durkan2019, Muller2019}. Instead of training
 models with a likelihood loss, in~\citep{Nash2019}, the authors use
 auto-regressive energy models for density estimation. While most methods either
 rely on coupling~\citep{Dinh2014, Dinh2016} or
 auto-regressive~\citep{Papamakarios2017, Kingma2016} models, transformation
 networks~\cite{Oliva2018} use a combination of both. The neural auto-regressive
 flows~\citep{Huang2018}, which have monotonic function guarantees, can be used
 as a drop-in replacement for the spline-based copula network used in this
 paper.

\section{Experiments}
\label{sec:experiemnts}
\begin{figure}
	\centering
	\includegraphics[width=\textwidth]{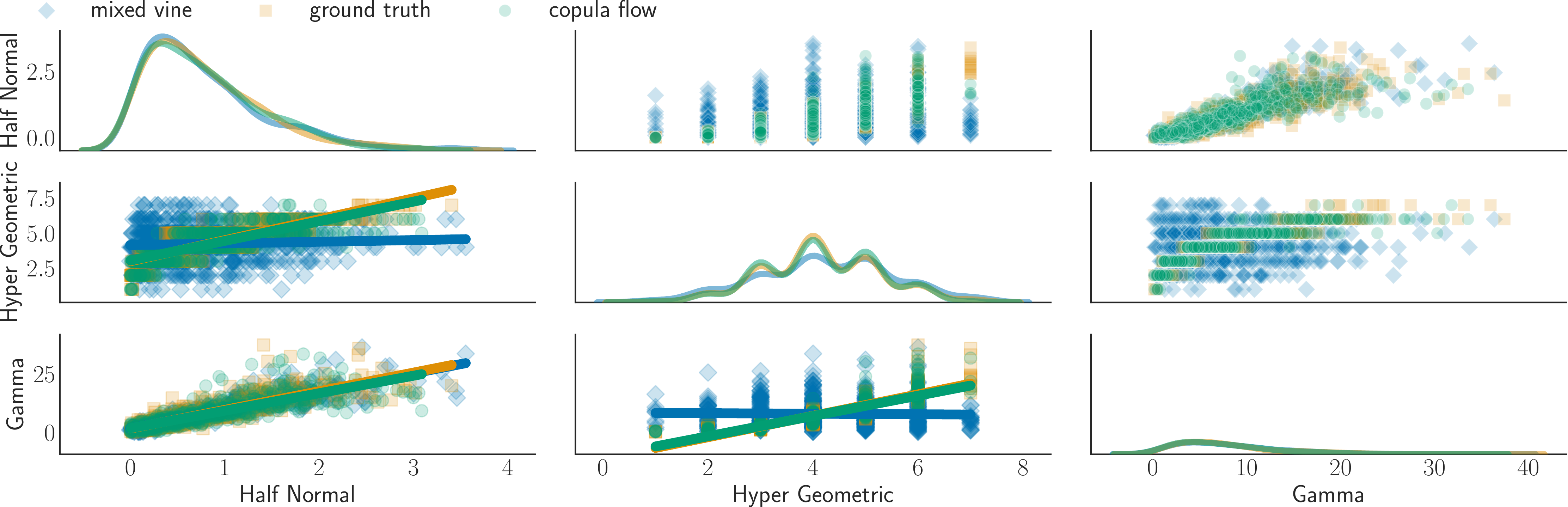}
	\caption{Mixed variable copula learning: joint marginals, joint scatter plots, regression plots. Fitting a mixed type vine~\citep{Onken2016}. $ X_2 $ is a
		discrete Hyper geometric, $ X_1 $ is Half Normal, and $X_3  $ is Gamma
		distributed. Pairwise copulas $\left( X_1 \text{ and } X_2 \right)$
		Gaussian, $\left( X_2 \text{ and } X_3 \right)$ Clayton, $\left( X_1
		\text{ and } X_3 \right)$ Gumbel. The diagonal plots show the marginals for each
		variable, whereas the upper triangle shows joint scatter plots. The lower triangle shows
		regression plots for each variable. Here, we overlay a
		linear regression fit and data for the pair of variables.  For continuous variables $X_1$ and $X_3$ all
		three methods produce regression lines similar to each other. For
		relations between discrete and continuous variables, the copula flow and the
		ground truth align. However, for both discrete to continuous relations, the
		mixed vine fails to capture the relation as evident by the misaligned
		linear regression fit. }
	\label{fig:mixedExperiment}
\end{figure}

Our proposed copula flow can learn copula models with continuous and discrete
variables. This allows us to use the model to learn complex relations in tabular
data as well.

\paragraph{Mixed Variable Copula Learning}
We start with an
illustrative example by constructing a generative model with discrete and
continuous variables. We demonstrate that the copula flow model can indeed learn the structure
as well as the marginal densities. We construct a vine model with three different
marginals: 1) $ X_1$, continuous,  Half Normal; 2) $X_2$ discrete,
hypergeometric; 3) $X_3$, continuous, Gamma. Since the copula describes a symmetric
relationship, we have three different bivariate relations between three
variables. We assign three widely used bivariate copulas: a Gaussian for $X_1$ and $X_2$, a Clayton for $X_2$ and $X_3$, and a Gumbel copula for  $X_1$ and $X_3$. This model is inspired by the mixed vine data model described 
in~\citep{Onken2016}, where authors propose to build a likelihood using CDFs and vine for mixed data. We use this model as a baseline for comparison.
The copula flow can  learn all three marginals, which are plotted
along the diagonal of Figure~\ref{fig:mixedExperiment}, while the mixed vine copula struggles a bit (second row, second column). The copula flow also successfully learns copulas between discrete and continuous variables, as evident by the linear regression line in the lower half of the plots in 
Figure~\ref{fig:mixedExperiment}. Both the 
mixed vine model~\citep{Onken2016} and the copula flow can learn the continuous copula well as evident by the left lowermost regression plot, but the mixed vine copula struggles with mixed variable copulas (third row, second column of Figure~\ref{fig:mixedExperiment}). Scatter plots in the upper diagonal show that the samples from mixed vine model are more spread out while copula flow samples align with the true samples.


\paragraph{Density Estimation}
For density estimation, we compare the performance against the current state of
the art neural density estimators. The standard benchmark introduced by
\citep{Papamakarios2017} modifies the original data with discrete marginals. For example, in the UCI power data-set, the authors
add a uniform random noise to the sub-metering columns. This is equivalent to using the
distributional transform, Definition~\ref{defn:dist_xform}, with a fixed seed.
These ad-hoc changes defeat the purpose of interpretability and transparency.
However, to maintain the exact comparison across different density estimators we
keep the data, training, test, and validation splits exactly as 
in~\citep{Papamakarios2017}, with the code provided by~\citep{Oliva2018}. We use
CDF splines flow for estimating the marginals, which is equivalent to estimating
the model with an independence copula.
We obtain the univariate uniform marginals by inverting the flow, and then
fit a copula flow model to obtain the joint density. 
 We summarise the density
estimation results in  Table~\ref{tbl:density_estim}. The marginal likelihood scores ignore the
relationships amongst the variables, whereas the copula likelihood shows information
gained by using copula. The joint likelihood can
be expressed as sum of marginal and copula likelihoods, see~\eqref{eqn:Loglikelihood}  The copula flow based
model achieves a performance that is close to the state of the art. Further
improvements may be achieved by fine tuning the copula flow structure as the
marginal CDFs are close to the empirical CDF as verified by using the Kolmogorov-Smirnov (KS) test; see Appendix~\ref{sec:extra_results}
 
 \begin{table}[th]
	\centering
	\begin{tabular}{c c c c c c }
		\hline 
		Model & Power  & Gas & Hepmass & Miniboone  \\ 
		\hline 
		FFJORD & $ 0.46 \pm 0.01 $ & $  8.59 \pm 0.12 $ & $ - 14.92 \pm 0.08 $ &
		$ - 10.43 \pm 0.04 $  \\
		RQ-NSF (AR)~\citep{Durkan2019} & $ 0.66 \pm 0.01 $ & $ 13.09 \pm 0.02 $ &  $ - 14.01 \pm
		0.03 $ & $ - 9.22 \pm 0.48 $  \\
		MAF~\citep{Papamakarios2017} &  $ 0.45 \pm 0.01 $ & $12.35 \pm 0.02$ & $-17.03 \pm 0.02$ &$-10.92
		\pm 0.46$   \\ 
		\hline
		Marginal Flows $ \flow $& $ -0.80 \pm 0.02 $ &  $ -6.67 \pm 0.02  $&  $
		- 26.42  \pm 0.05$
		& $-53.17 \pm 0.06$  \\ 
		Copula Flow $ \copulaflow	 $& $ 1.39 \pm 0.03 $ & $ 15.6 \pm 0.67$ &
		$5.4 \pm 0.10$ & $37.77 \pm 0.21 $  \\ 
		Joint Model $ \copulaflow	\, + \,  \flow  $ & $ 0.59 \pm 0.03 $
		&  $ 8.05 \pm 0.68 $ & $-19.6 \pm 0.12$ & $-14.83 \pm 0.21$  \\ 
		\hline 
	\end{tabular} 
	\caption{Test log-likelihoods (nats) on density estimation benchmarks. 
	Exact replication of the dataset and splits as provided
	by~\citep{Papamakarios2017}.}
	\label{tbl:density_estim}
\end{table}


\paragraph{Synthetic Data Generation}
 Estimating the performance of data generators is a challenging
 process~\citep{Theis2016}. Merely having very high likelihood on the training
 or test set is not necessarily the best metric. One measure to look at is the
 machine learning efficacy of the generated data, i.e., whether can we use the synthetic
 data to effectively perform machine learning tasks (e.g., regression or
 classification), which we wished to accomplish with the true
 data~\citep{Xu2019, Jordon2019}.  Here, we test the synthetic data generator
 using the framework introduced in~\citep{Xu2019}.
 The copula flow model can capture the relations between different
 variables while learning a density model as it is evident by the high scores in the
 ML tasks benchmark; see
 Table~\ref{tbl:synth_data}. For Gaussian mixture (GM) based models, the copula flow model performs very close to the `Identity'
 transform, which uses the true data. This indicates a high similarity of the true and the synthetic data-set. For real data, the model
 performs very close to the current state of the art~\citep{Xu2019} for
 classification tasks and outperforms it on regression tasks on aggregate. The
 quality of learned CDFs is assessed by using the KS test; see Appendix~\ref{sec:extra_results}.
 
\begin{table}[ht] 
	\centering
\begin{tabular}{c c c c c c c c c c c}
\hline
                    & \multicolumn{2}{c}{GM Sim.}                &  &  & \multicolumn{2}{c}{BN Sim.}                &  &  & \multicolumn{2}{c}{Real} \\ \cline{2-3} \cline{6-7} \cline{10-11} 
Method              & $\mathcal{L}_{syn}$ & $\mathcal{L}_{test}$ &  &  & $\mathcal{L}_{syn}$ & $\mathcal{L}_{test}$ &  &  & clf             & reg    \\ \hline
Identity                      & -2.61  & -2.61          &  &  & -9.33  & -9.36  &  &  & 0.743 & 0.14           \\ \hline
CLBN~\citep{Che2017}          & -3.06  & 7.31           &  &  & -10.66 & -9.92  &  &  & 0.382 & -6.28          \\
PrivBN~\citep{Zhang2017}      & -3.38  & -12.42         &  &  & -12.97 & -10.90 &  &  & 0.225 & -4.49          \\
MedGAN~\citep{Choi2017}       & -7.27  & -60.03         &  &  & -11.14 & -12.15 &  &  & 0.137 & -8.80          \\
VeeGAN~\citep{Srivastava2017} & -10.06 & -4.22          &  &  & -15.40 & -13.86 &  &  & 0.143 & -6.5e6         \\
TableGAN~\citep{Park2018}     & -8.24  & -4.12          &  &  & -11.84 & -10.47 &  &  & 0.162 & -3.09          \\
TVAE~\citep{Xu2019} & \textbf{-2.65}      & -5.42                &  &  & \textbf{-6.76}      & \textbf{-9.59}       &  &  & \textbf{0.519}  & -0.20  \\
CTGAN~\citep{Xu2019}          & -5.72  & -3.40          &  &  & -11.67 & -10.60 &  &  & 0.469 & -0.43          \\ \hline
CopulaFlow                    & -3.17  & \textbf{-2.98} &  &  & -10.51 & -9.69  &  &  & 0.431 & \textbf{-0.18} \\ \hline
\end{tabular}
\caption{Performance of the copula flow models for synthetic data generation. 
We benchmark by using the synthetic data to perform the 
machine learning (ML) task, classification or regression, associated with
the data-set.  The metric is generated for the original test data.
Gaussian mixture (GM) and Bayesian network (BN) simulations correspond
to data generated from known GM or BN models. Log-likelihoods $\mathcal{L}_{syn}$
of  the true model given synthetic data and $\mathcal{L}_{test}$ of the 
generative model given the test data are shown. All values except the copula 
flow are from the framework in~\citep{Xu2019}.  }
	\label{tbl:synth_data}
\end{table}
%

\section{Conclusion}
In this work we proposed a new probabilistic model to generate high fidelity
synthetic data. Our model is based on copula theory which allows us to build
interpretable and flexible model for the data generation. We used distributional
transform to address the challenges associated with learning copulas for
discrete data. Experimental results show that our proposed model is capable of
learning mixed variable copulas required for tabular data-sets. We show that
synthetic data generated from a variety of real data-sets is capable of capturing
the relationship amongst the mixed variables.
Our model can be extended to generate data with differential
privacy guarantees. We consider this as an important extension and future work.
Moreover, our copula flow model can also be used, where copula models are used,
e.g., in finance and actuarial science.

\section*{Broader Impact}
 Copula based density models are interpretable, as combination of individual
behaviour and their joint tendencies independent of individual behaviour. Hence,
these models are used in financial and insurance modelling, where interpretable
models are needed for regulatory purposes. Appeal for synthetic data stems from
inherent data privacy achieved by the model as it hides true data. However, the
current implementation is still prone to privacy attacks as one can use a large
number of queries to learn the statistics from the model, i.e., private data may
leak through the synthetic data generated by the proposed model. Copula models
are typically governed by a small number of parameters, whereas the copula
flow model has very large deep network. This can pose a challenge even if the
output of the network is interpretable.


\bibliography{copulaMAF}

\begin{thebibliography}{48}
\providecommand{\natexlab}[1]{#1}
\providecommand{\url}[1]{\texttt{#1}}
\expandafter\ifx\csname urlstyle\endcsname\relax
  \providecommand{\doi}[1]{doi: #1}\else
  \providecommand{\doi}{doi: \begingroup \urlstyle{rm}\Url}\fi

\bibitem[Aas et~al.(2009)Aas, Czado, Frigessi, and Bakken]{Aas2009}
K.~Aas, C.~Czado, A.~Frigessi, and H.~Bakken.
\newblock {Pair-Copula Constructions of Multiple Dependence}.
\newblock \emph{Insurance: Mathematics and Economics}, 44\penalty0
  (2):\penalty0 182--198, apr 2009.
\newblock ISSN 01676687.
\newblock \doi{10.1016/j.insmatheco.2007.02.001}.
\newblock URL
  \url{http://www.sciencedirect.com/science/article/pii/S0167668707000194}.

\bibitem[Arjovsky and Bottou(2017)]{Arjovsky2017}
M.~Arjovsky and L.~Bottou.
\newblock {Towards Principled Methods for Training Generative Adversarial
  Networks}.
\newblock In \emph{International Conference on Learning Representations, ICLR},
  2017.
\newblock URL \url{http://arxiv.org/abs/1701.04862}.

\bibitem[Beaulieu-Jones et~al.(2019)Beaulieu-Jones, Wu, Williams, Lee,
  Bhavnani, Byrd, and Greene]{Beaulieu2019}
B.~K. Beaulieu-Jones, Z.~S. Wu, C.~Williams, R.~Lee, S.~P. Bhavnani, J.~B.
  Byrd, and C.~S. Greene.
\newblock {Privacy-Preserving Generative Deep Neural Networks Support Clinical
  Data Sharing}.
\newblock \emph{Circulation. Cardiovascular quality and outcomes}, 2019.
\newblock ISSN 19417705.
\newblock \doi{10.1161/CIRCOUTCOMES.118.005122}.
\newblock URL
  \url{https://www.ahajournals.org/doi/10.1161/CIRCOUTCOMES.118.005122
  http://www.ncbi.nlm.nih.gov/pubmed/31284738}.

\bibitem[Chang et~al.(2019)Chang, Pan, and Joe]{Chang2019}
B.~Chang, S.~Pan, and H.~Joe.
\newblock {Vine Copula Structure Learning via Monte Carlo Tree Search}.
\newblock In \emph{Proceedings of Machine Learning Research}, 2019.
\newblock URL \url{http://proceedings.mlr.press/v89/chang19a.html}.

\bibitem[Che et~al.(2017)Che, Cheng, Zhai, Sun, and Liu]{Che2017}
Z.~Che, Y.~Cheng, S.~Zhai, Z.~Sun, and Y.~Liu.
\newblock {Boosting Deep Learning Risk Prediction with Generative Adversarial
  Networks for Electronic Health Records}.
\newblock In \emph{Proceedings - IEEE International Conference on Data Mining,
  ICDM}, 2017.
\newblock ISBN 9781538638347.
\newblock \doi{10.1109/ICDM.2017.93}.
\newblock URL \url{http://arxiv.org/abs/1709.01648}.

\bibitem[Choi et~al.(2017)Choi, Biswal, Malin, Duke, Stewart, and
  Sun]{Choi2017}
E.~Choi, S.~Biswal, B.~Malin, J.~Duke, W.~F. Stewart, and J.~Sun.
\newblock {Generating Multi-label Discrete Patient Records using Generative
  Adversarial Networks}.
\newblock In \emph{Machine Learning in Health Care (MLHC)}, mar 2017.
\newblock URL \url{http://arxiv.org/abs/1703.06490}.

\bibitem[Czado(2010)]{Czado2010}
C.~Czado.
\newblock {Pair-Copula Constructions of Multivariate Copulas BT - Copula Theory
  and Its Applications}.
\newblock In \emph{Copula Theory and Its Applications}, volume 198, pages
  93--109. Springer, 2010.
\newblock ISBN 978-3-642-12465-5.
\newblock \doi{10.1007/978-3-642-12465-5}.
\newblock URL \url{https://www.jstor.org/stable/3663514
  http://www.springerlink.com/index/10.1007/978-3-642-12465-5{\_}4{\%}5Cnpapers2://publication/doi/10.1007/978-3-642-12465-5{\_}4}.

\bibitem[David and Johnson(1948)]{David1948}
F.~N. David and N.~L. Johnson.
\newblock The probability integral transformation when parameters are estimated
  from the sample.
\newblock \emph{Biometrika}, 35\penalty0 (1/2):\penalty0 182--190, 1948.
\newblock ISSN 00063444.
\newblock URL \url{http://www.jstor.org/stable/2332638}.

\bibitem[Dillon et~al.(2017)Dillon, Langmore, Tran, Brevdo, Vasudevan, Moore,
  Patton, Alemi, Hoffman, and Saurous]{Dillon2017}
J.~V. Dillon, I.~Langmore, D.~Tran, E.~Brevdo, S.~Vasudevan, D.~Moore,
  B.~Patton, A.~Alemi, M.~Hoffman, and R.~A. Saurous.
\newblock {TensorFlow Distributions}.
\newblock \emph{ArXiv Preprint}, nov 2017.
\newblock URL \url{http://arxiv.org/abs/1711.10604}.

\bibitem[Dinh et~al.(2015)Dinh, Krueger, and Bengio]{Dinh2014}
L.~Dinh, D.~Krueger, and Y.~Bengio.
\newblock {NICE: Non-linear Independent Components Estimation}.
\newblock In \emph{3rd International Conference on Learning Representations,
  ICLR 2015 - Workshop Track Proceedings}, 2015.
\newblock URL \url{http://arxiv.org/abs/1410.8516}.

\bibitem[Dinh et~al.(2019)Dinh, Sohl-Dickstein, and Bengio]{Dinh2016}
L.~Dinh, J.~Sohl-Dickstein, and S.~Bengio.
\newblock {Density Estimation using Real NVP}.
\newblock In \emph{5th International Conference on Learning Representations,
  ICLR}, 2019.
\newblock URL \url{http://arxiv.org/abs/1605.08803}.

\bibitem[Durkan et~al.(2019)Durkan, Bekasov, Murray, and
  Papamakarios]{Durkan2019}
C.~Durkan, A.~Bekasov, I.~Murray, and G.~Papamakarios.
\newblock {Neural Spline Flows}.
\newblock In \emph{Advances in Neural Information Processing Systems},
  Vancouver, 2019.
\newblock URL \url{http://arxiv.org/abs/1906.04032}.

\bibitem[Dwork and Roth(2013)]{Dwork2013}
C.~Dwork and A.~Roth.
\newblock {The Algorithmic Foundations of Differential Privacy}.
\newblock \emph{Foundations and Trends in Theoretical Computer Science}, 2013.
\newblock ISSN 15513068.
\newblock \doi{10.1561/0400000042}.

\bibitem[Elidan(2010)]{Elidan2010}
G.~Elidan.
\newblock {Copula Bayesian Networks}.
\newblock In \emph{Advances in Neural Information Processing Systems}, 2010.
\newblock URL \url{http://papers.nips.cc/paper/3956-copula-bayesian-networks}.

\bibitem[Elliot and Domingo-Ferrer(2018)]{Elliot2018}
M.~Elliot and J.~Domingo-Ferrer.
\newblock {The Future of Statistical Disclosure Control}.
\newblock Technical report, Goverment Statistical Service, London, UK, dec
  2018.
\newblock URL \url{http://arxiv.org/abs/1812.09204}.

\bibitem[Germain et~al.(2015)Germain, Gregor, Murray, and
  Larochelle]{Germain2015}
M.~Germain, K.~Gregor, I.~Murray, and H.~Larochelle.
\newblock {MADE: Masked Autoencoder for Distribution Estimation}.
\newblock \emph{32nd International Conference on Machine Learning, ICML}, feb
  2015.
\newblock URL \url{http://arxiv.org/abs/1502.03509}.

\bibitem[Goodfellow et~al.(2014)Goodfellow, Pouget-Abadie, Mirza, Xu,
  Warde-Farley, Ozair, Courville, and Bengio]{Goodfellow2014}
I.~J. Goodfellow, J.~Pouget-Abadie, M.~Mirza, B.~Xu, D.~Warde-Farley, S.~Ozair,
  A.~Courville, and Y.~Bengio.
\newblock {Generative Adversarial Nets}.
\newblock In \emph{Advances in Neural Information Processing Systems}, 2014.
\newblock \doi{10.3156/jsoft.29.5_177_2}.
\newblock URL \url{http://arxiv.org/abs/1406.2661}.

\bibitem[Gregory and Delbourgo(1982)]{Gregory1982}
J.~A. Gregory and R.~Delbourgo.
\newblock {Piecewise Rational Quadratic Interpolation to Monotonic Data}.
\newblock \emph{IMA Journal of Numerical Analysis}, 2\penalty0 (2):\penalty0
  123--130, apr 1982.
\newblock ISSN 02724979.
\newblock \doi{10.1093/imanum/2.2.123}.
\newblock URL
  \url{https://academic.oup.com/imajna/article-lookup/doi/10.1093/imanum/2.2.123}.

\bibitem[Hansen(1994)]{Hansen1994}
B.~E. Hansen.
\newblock {Autoregressive Conditional Density Estimation}.
\newblock \emph{International Economic Review}, 35\penalty0 (3):\penalty0 705,
  aug 1994.
\newblock ISSN 00206598.
\newblock \doi{10.2307/2527081}.
\newblock URL \url{https://www.jstor.org/stable/2527081?origin=crossref}.

\bibitem[Huang et~al.(2018)Huang, Krueger, Lacoste, and Courville]{Huang2018}
C.~W. Huang, D.~Krueger, A.~Lacoste, and A.~Courville.
\newblock {Neural Autoregressive Flows}.
\newblock In \emph{35th International Conference on Machine Learning, ICML},
  apr 2018.
\newblock ISBN 9781510867963.
\newblock URL \url{http://arxiv.org/abs/1804.00779}.

\bibitem[Hyv{\"{a}}rinen and Pajunen(1999)]{Hyvarinen1999}
A.~Hyv{\"{a}}rinen and P.~Pajunen.
\newblock {Nonlinear Independent Component Analysis: Existence and Uniqueness
  Results}.
\newblock \emph{Neural Networks}, 12\penalty0 (3):\penalty0 429--439, apr 1999.
\newblock ISSN 08936080.
\newblock \doi{10.1016/S0893-6080(98)00140-3}.
\newblock URL
  \url{https://www.sciencedirect.com/science/article/abs/pii/S0893608098001403}.

\bibitem[Joe(2005)]{Joe2005}
H.~Joe.
\newblock {Asymptotic Efficiency of the Two-stage Estimation Method for
  Copula-based Models}.
\newblock \emph{Journal of Multivariate Analysis}, 94\penalty0 (2):\penalty0
  401--419, jun 2005.
\newblock ISSN 0047-259X.
\newblock \doi{10.1016/J.JMVA.2004.06.003}.
\newblock URL
  \url{https://www.sciencedirect.com/science/article/pii/S0047259X04001289}.

\bibitem[Jordon et~al.(2019)Jordon, Yoon, and {Van Der Schaar}]{Jordon2019}
J.~Jordon, J.~Yoon, and M.~{Van Der Schaar}.
\newblock {PATE-GaN: Generating synthetic data with differential privacy
  guarantees}.
\newblock In \emph{7th International Conference on Learning Representations,
  ICLR}, 2019.
\newblock URL
  \url{https://www.semanticscholar.org/paper/PATE-GAN{\%}3A-Generating-Synthetic-Data-with-Privacy-Jordon-Yoon/af1841e1db6579f1f1777a59c7e9e4658d2ac466}.

\bibitem[Kingma and Welling(2013)]{Kingma2013}
D.~P. Kingma and M.~Welling.
\newblock {Auto-Encoding Variational Bayes}.
\newblock \emph{arXiv preprint}, 2013.
\newblock URL \url{http://arxiv.org/abs/1312.6114}.

\bibitem[Kingma et~al.(2016)Kingma, Salimans, Jozefowicz, Chen, Sutskever, and
  Welling]{Kingma2016}
D.~P. Kingma, T.~Salimans, R.~Jozefowicz, X.~Chen, I.~Sutskever, and
  M.~Welling.
\newblock {Improved Variational Inference with Inverse Autoregressive Flow
  Diederik}, jun 2016.
\newblock URL \url{https://arxiv.org/pdf/1606.04934.pdf
  http://arxiv.org/abs/1606.04934}.

\bibitem[Lopez-paz et~al.(2013)Lopez-paz, Hern{\'{a}}ndez-lobato, and
  Ghahramani]{Lopez-paz2013}
D.~Lopez-paz, J.~M. Hern{\'{a}}ndez-lobato, and Z.~Ghahramani.
\newblock {Gaussian Process Vine Copulas for Multivariate Dependence}.
\newblock In \emph{30th International Conference on Machine Learning, ICML},
  2013.
\newblock URL
  \url{http://machinelearning.wustl.edu/mlpapers/papers/lopez-paz13}.

\bibitem[M{\"{u}}ller et~al.(2019)M{\"{u}}ller, Mcwilliams, Rousselle, Gross,
  and Nov{\'{a}}k]{Muller2019}
T.~M{\"{u}}ller, B.~Mcwilliams, F.~Rousselle, M.~Gross, and J.~Nov{\'{a}}k.
\newblock {Neural Importance Sampling}.
\newblock \emph{ACM Transactions on Graphics}, 38\penalty0 (5):\penalty0 1--19,
  oct 2019.
\newblock ISSN 07300301.
\newblock \doi{10.1145/3341156}.
\newblock URL \url{http://arxiv.org/abs/1808.03856
  http://arxiv.org/abs/1907.00503
  http://dl.acm.org/citation.cfm?doid=3341165.3341156}.

\bibitem[Nash and Durkan(2019)]{Nash2019}
C.~Nash and C.~Durkan.
\newblock {Autoregressive Energy Machines}.
\newblock \emph{ArXiv Preprint}, apr 2019.
\newblock URL \url{http://arxiv.org/abs/1904.05626}.

\bibitem[Nelsen(2006)]{Nelsen2006}
R.~B. Nelsen.
\newblock \emph{{An Introduction to Copulas}}.
\newblock Springer New York, New York, NY, 2006.
\newblock ISBN 978-0-387-28678-5.
\newblock \doi{10.1007/0-387-28678-0_1}.
\newblock URL \url{http://link.springer.com/10.1007/0-387-28678-0{\_}1}.

\bibitem[Nikoloulopoulos and Karlis(2009)]{Nikoloulopoulos2009}
A.~K. Nikoloulopoulos and D.~Karlis.
\newblock {Modeling Multivariate Count Data Using Copulas}.
\newblock \emph{Communications in Statistics - Simulation and Computation},
  39\penalty0 (1):\penalty0 172--187, dec 2009.
\newblock ISSN 0361-0918.
\newblock \doi{10.1080/03610910903391262}.
\newblock URL
  \url{http://www.tandfonline.com/doi/abs/10.1080/03610910903391262}.

\bibitem[Oliva et~al.(2018)Oliva, Dubey, Zaheer, P{\'{o}}czos, Salakhutdinov,
  Xing, and Schneider]{Oliva2018}
J.~B. Oliva, A.~Dubey, M.~Zaheer, B.~P{\'{o}}czos, R.~Salakhutdinov, E.~P.
  Xing, and J.~Schneider.
\newblock {Transformation Autoregressive Networks}.
\newblock In \emph{35th International Conference on Machine Learning, ICML},
  volume~9, 2018.
\newblock ISBN 9781510867963.
\newblock URL \url{http://arxiv.org/abs/1801.09819}.

\bibitem[Onken and Panzeri(2016)]{Onken2016}
A.~Onken and S.~Panzeri.
\newblock {Mixed Vine Copulas as Joint Models of Spike Counts and Local Field
  Potentials}.
\newblock In \emph{Advances in Neural Information Processing Systems}, 2016.
\newblock URL
  \url{http://papers.nips.cc/paper/6068-mixed-vine-copulas-as-joint-models-of-spike-counts-and-local-field-potentials}.

\bibitem[Papamakarios et~al.(2017)Papamakarios, Pavlakou, and
  Murray]{Papamakarios2017}
G.~Papamakarios, T.~Pavlakou, and I.~Murray.
\newblock {Masked Autoregressive Flow for Density Estimation}.
\newblock \emph{Advances in Neural Information Processing Systems}, may 2017.
\newblock ISSN 10495258.
\newblock URL \url{http://arxiv.org/abs/1705.07057}.

\bibitem[Papamakarios et~al.(2019)Papamakarios, Nalisnick, Rezende, Mohamed,
  and Lakshminarayanan]{Papamakarios2019}
G.~Papamakarios, E.~Nalisnick, D.~J. Rezende, S.~Mohamed, and
  B.~Lakshminarayanan.
\newblock {Normalizing Flows for Probabilistic Modeling and Inference}.
\newblock \emph{ArXiv Preprint}, dec 2019.
\newblock URL \url{https://arxiv.org/abs/1912.02762
  http://arxiv.org/abs/1912.02762}.

\bibitem[Park et~al.(2018)Park, Mohammadi, Gorde, Jajodia, Park, and
  Kim]{Park2018}
N.~Park, M.~Mohammadi, K.~Gorde, S.~Jajodia, H.~Park, and Y.~Kim.
\newblock {Data Synthesis based on Generative Adversarial Networks}.
\newblock In \emph{Proceedings of the VLDB Endowment}, jun 2018.
\newblock \doi{10.14778/3231751.3231757}.
\newblock URL \url{http://arxiv.org/abs/1806.03384
  http://dx.doi.org/10.14778/3231751.3231757}.

\bibitem[Patki et~al.(2016)Patki, Wedge, and Veeramachaneni]{Patki2016}
N.~Patki, R.~Wedge, and K.~Veeramachaneni.
\newblock {The Synthetic Data Vault}.
\newblock In \emph{3rd IEEE International Conference on Data Science and
  Advanced Analytics, DSAA}, 2016.
\newblock ISBN 9781509052066.
\newblock \doi{10.1109/DSAA.2016.49}.

\bibitem[Rezende and Mohamed(2015)]{Rezende2015}
D.~J. Rezende and S.~Mohamed.
\newblock {Variational Inference with Normalizing Flows}.
\newblock In \emph{32nd International Conference on Machine Learning, ICML},
  2015.
\newblock ISBN 9781510810587.
\newblock URL \url{https://arxiv.org/pdf/1505.05770.pdf}.

\bibitem[R{\"{u}}schendorf(2009)]{Ruschendorf2009}
L.~R{\"{u}}schendorf.
\newblock {On the Distributional Transform, Sklar's Theorem, and the Empirical
  Copula Process}.
\newblock \emph{Journal of Statistical Planning and Inference}, 2009.
\newblock ISSN 03783758.
\newblock \doi{10.1016/j.jspi.2009.05.030}.

\bibitem[Salimans et~al.(2019)Salimans, Karpathy, Chen, and
  Kingma]{Salimans2019}
T.~Salimans, A.~Karpathy, X.~Chen, and D.~P. Kingma.
\newblock {PixelCNN++: Improving the PixelCNN with Discretized Logistic Mixture
  Likelihood and Other Modifications}.
\newblock In \emph{5th International Conference on Learning Representations,
  ICLR}, 2019.
\newblock URL \url{http://arxiv.org/abs/1701.05517}.

\bibitem[Sklar(1959)]{Sklar1959}
M.~Sklar.
\newblock {Fonctions de R{\'{e}}partition {\`{a}} n Dimensions Et Leurs
  Marges}.
\newblock \emph{Publications of Institute Statistics University Paris (in
  French) 8}, 1959.
\newblock URL
  \url{https://scholar.google.co.uk/scholar?hl=en{\&}q={\%}22Fonctions+de+r{\'{e}}partition+{\`{a}}+n+dimensions+et+leurs+marges{\%}22{\&}btnG={\&}as{\_}sdt=1{\%}2C5{\&}as{\_}sdtp={\#}0}.

\bibitem[Srivastava et~al.(2017)Srivastava, Valkov, Russell, Gutmann, and
  Sutton]{Srivastava2017}
A.~Srivastava, L.~Valkov, C.~Russell, M.~U. Gutmann, and C.~Sutton.
\newblock {VEEGAN: Reducing Mode Collapse in GANs using Implicit Variational
  Learning}.
\newblock In \emph{Advances in Neural Information Processing Systems}, 2017.

\bibitem[Tabak and Vanden-Eijnden(2010)]{Tabak2010}
E.~G. Tabak and E.~Vanden-Eijnden.
\newblock {Density Estimation by Dual Ascent of the Log-likelihood}.
\newblock \emph{Communications in Mathematical Sciences}, 2010.
\newblock ISSN 15396746.
\newblock \doi{10.4310/CMS.2010.v8.n1.a11}.

\bibitem[Theis et~al.(2016)Theis, {Van Den Oord}, and Bethge]{Theis2016}
L.~Theis, A.~{Van Den Oord}, and M.~Bethge.
\newblock {A Note on the Evaluation of Generative Models}.
\newblock In \emph{4th International Conference on Learning Representations,
  ICLR}, 2016.
\newblock URL \url{http://arxiv.org/abs/1511.01844}.

\bibitem[Tran et~al.(2019)Tran, Vafa, Agrawal, Dinh, and Poole]{Tran2019}
D.~Tran, K.~Vafa, K.~K. Agrawal, L.~Dinh, and B.~Poole.
\newblock {Discrete flows: Invertible Generative Models of Discrete ata}.
\newblock In \emph{Deep Generative Models for Highly Structured Data, DGS@ICLR
  2019 Workshop}, 2019.
\newblock URL \url{http://arxiv.org/abs/1905.10347}.

\bibitem[Wang et~al.(2019)Wang, She, and Ward]{wang2019}
Z.~Wang, Q.~She, and T.~E. Ward.
\newblock Generative adversarial networks: A survey and taxonomy.
\newblock \emph{arXiv preprint}, 2019.

\bibitem[Xu et~al.(2019)Xu, Skoularidou, Cuesta-Infante, and
  Veeramachaneni]{Xu2019}
L.~Xu, M.~Skoularidou, A.~Cuesta-Infante, and K.~Veeramachaneni.
\newblock {Modeling Tabular data using Conditional GAN}.
\newblock In \emph{Advances in Neural Information Processing Systems}, jun
  2019.
\newblock URL \url{http://arxiv.org/abs/1907.00503}.

\bibitem[Zhang et~al.(2017)Zhang, Cormode, Procopiuc, Srivastava, and
  Xiao]{Zhang2017}
J.~Zhang, G.~Cormode, C.~M. Procopiuc, D.~Srivastava, and X.~Xiao.
\newblock {Priv bayes: Private Data Release via Bayesian Networks}.
\newblock \emph{ACM Transactions on Database Systems}, 2017.
\newblock ISSN 15574644.
\newblock \doi{10.1145/3134428}.

\bibitem[Ziegler and Rush(2019)]{Ziegler2019}
Z.~M. Ziegler and A.~M. Rush.
\newblock {Latent Normalizing Flows for Discrete Sequences}.
\newblock In \emph{36th International Conference on Machine Learning, ICML},
  2019.
\newblock ISBN 9781510886988.
\newblock URL \url{http://arxiv.org/abs/1901.10548}.

\end{thebibliography}

\onecolumn
\newpage
\appendix
\section{Flow Model}
\label{sec:apendixFlow}

We can derive the flow-based likelihood by starting from the generative procedure
for the samples. We start with a $d$-dimensional independently distributed random
vector $\vec U \sim \uniform{1}^d$ and map it through the copula flow $\copulaflow_{\vec
X}$ to obtain the random vector $\vec U_X  = \copulaflow_{\vec X} (\vec U)$. The
joint vector is then mapped through the marginal flows $\vec \flow_{\vec X}$ to
obtain a random vector $\vec X = \flow_{\vec X} (\vec U_X)$. The combined
formulation can be written as a composition of two flows to obtain $\vec X =
\flow_{\vec X} \left( \copulaflow_{\vec X} (\vec U) \right)$, which is also a
valid flow function~\citep{Rezende2015}.

The likelihood for this flow function can be written as
\begin{align}
\label{eqn:likelihood_part1}
	\vec f_{\vec X} (\vec X)  &= \vec f_{\vec U_{\vec X}}(\vec \flow_{\vec X}\inv (\vec X)) \, \left| \mathrm{det} \left(\frac{\partial  \vec \flow_{\vec X}\inv(\vec X)}{\partial \vec X} \right) \right|. 
\end{align}
The marginal flows are independent for each dimension of the random vector $\vec
X$. Hence, the determinant can be expressed as a product of the diagonal terms to
obtain the likelihood 
\begin{align}
	\vec f_{\vec X} (\vec X)  &= \vec f_{\vec U_{\vec X}}(\vec \flow_{\vec X}\inv (\vec X)) \, \prod_{i=1}^{d} \left| \left(\frac{\partial  \flow_{X_i}\inv(x_i)}{\partial x_i} \right) \right|.
\end{align}
The quantity $\vec f_{\vec U_{\vec X}}(\vec \flow_{\vec X}\inv (\vec X))$ is
essentially the flow likelihood for the copula density, which we can write as 
\begin{align}
	\vec f_{\vec U_{\vec X}}(\vec \flow_{\vec X}\inv (\vec X)) & = \vec f_{\vec U_{\vec X}} \left( \vec \copulaflow_{\vec X} (\vec U) \right) \\
	& = \vec f_{\vec U}(\vec \copulaflow_{\vec X}\inv (\vec U_{\vec X})) \,  \left|\mathrm{det} \left(\frac{\partial  \vec \copulaflow_{\vec X}\inv(\vec U_{\vec X})}{\partial \vec U_{\vec X}} \right) \right| \\
	& = \left| \mathrm{det} \left(\frac{\partial  \vec \copulaflow_{\vec X}\inv(\vec U_{\vec X})}{\partial \vec U_{\vec X}} \right) \right|,
	\label{eqn:copula_likelihood}
\end{align}
where $f_{\vec U}\left(\copulaflow_{\vec X}\inv (\vec U_{\vec X})\right) = f_{\vec U}(\vec U) = 1.0$, since $f_{\vec U}=\uniform{1}$.
Using the the copula likelihood in~\eqref{eqn:likelihood_part1} in
the total likelihood~\eqref{eqn:copula_likelihood} yields
\begin{align}
	\vec f_{\mat X}(\vec X) &= \left| \mathrm{det} \left(\frac{\partial  \vec \copulaflow_{\vec X}\inv(\vec \flow_{\vec X}\inv (\vec X))}{\partial \vec \flow_{\vec X}\inv (\vec X)} \right) \right| \prod_{i=1}^{d} \left| \left(\frac{\partial  \flow_{X_i}\inv(X_i)}{\partial X_i} \right) \right|.
\end{align}




\paragraph{Sklar's Theorem}
As discussed in section~\ref{sec:background}, we can interpret the inverse of the flow
function as a CDF: 
\begin{align}
	&\vec X = \flow_{\vec X} \left( \copulaflow_{\vec X} (\vec U) \right), \\
	&\copulaflow_{\vec X}\inv \left( \flow_{\vec X}\inv ( \vec X) \right)=  \vec U.
\end{align}
If we replace the flow function with the
true marginal CDF $\vec F_{\vec X}$ and the true copula CDF $\vec C_{\vec X}$, we can rewrite it as
\begin{align}
	\vec C_{\vec X} \left( \vec F_{\vec X}( \vec X) \right)&=  \vec U.
\end{align}
By the probability integral transform we know that if $\vec H_{\vec X}$ is the
CDF of $\vec X$ then $\vec H_{\vec X} (\vec X)= \vec U$. Hence, we can write the
joint CDF in terms of marginal CDFs as 
\begin{align}
	\vec H_{\vec X}( \vec X)  = \vec C_{\vec X} \left( \vec F_{\vec X}( \vec X) \right).
\end{align}
This is essentially Sklar's theorem~\citep{Sklar1959}. Note we
reached this result by  starting from the normalising flow formulation, whereas in
section~\ref{sec:background} we used Sklar's theorem for the generative model.

\section{Additional Results}
\label{sec:extra_results}
\subsection{Copula Fitting}
The majority of practical copula models are bivariate in nature and have the additional
advantage of easy visualisation. While the major theme of this work is ability
to learn copula for mixed variables types we demonstrate that the proposed
model can learn standard bivariate copulas easily. 

\subsection{Marginal fitting }

For arbitrarily complex models it is difficult to test the goodness of fit as the
likelihood only gives an indication for specific samples. Moreover, average likelihood values do not gives a reliable comparison metric. To compare samples generated from the mode we perform a 2 sample Kolmogorov-Smirnov (KS) test for goodness of fit. 

\begin{table}[]
\centering
\begin{tabular}{ccclcc}
\hline
\multicolumn{3}{c}{Power}                                          &  & \multicolumn{2}{c}{Gas} \\ \cline{1-3} \cline{5-6} 
                     & KS Stat              & P value              &  & KS Stat    & P value    \\ \cline{1-3} \cline{5-6} 
1                    & 0.012                & 0.474                &  & 0.009      & 0.755      \\
2                    & 0.011                & 0.582                &  & 0.017      & 0.099      \\
3                    & 0.012                & 0.521                &  & 0.017      & 0.39       \\
4                    & 0.014                & 0.320                &  & 0.016      & 0.509      \\
5                    & 0.012                & 0.452                &  & 0.014      & 0.256      \\
6                    & 0.009                & 0.792                &  & 0.016      & 0.148      \\
\multicolumn{1}{l}{} & \multicolumn{1}{l}{} & \multicolumn{1}{l}{} &  & 0.010      & 0.685      \\
\multicolumn{1}{l}{} & \multicolumn{1}{l}{} & \multicolumn{1}{l}{} &  & 0.011      & 0.590      \\ \hline
\end{tabular}
\caption{The table shows the Kolmogorov-Smirnov KS test for each of the six dimensions. We perform a two-sample KS test. A total of 10,000 samples were selected as the two-sample KS test is exact up to 10,000 data points; it uses approximations for large sample sizes. Low values of test statistic and high p-values indicates we cannot reject the hypothesis that two samples are from the same distribution.}
\label{tab:power-kstest}
\end{table}

\begin{figure*}[ht]
	\centering
	\subfloat[Marginal flow inverse $\vec U_{\vec X} = \flow_{\vec X}\inv (\vec X)$]{%
		\includegraphics[width=0.48\linewidth]{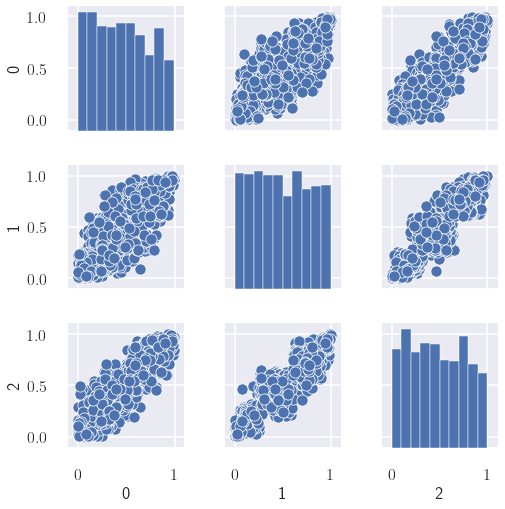}%
		\label{fig:marginal_inverse}%
	} 
	\subfloat[Copula flow inverse $\vec U = \copulaflow_{\vec U_{\vec X}}\inv ({\vec U_{\vec X}}) $ ]{%
		\includegraphics[width=0.48\linewidth]{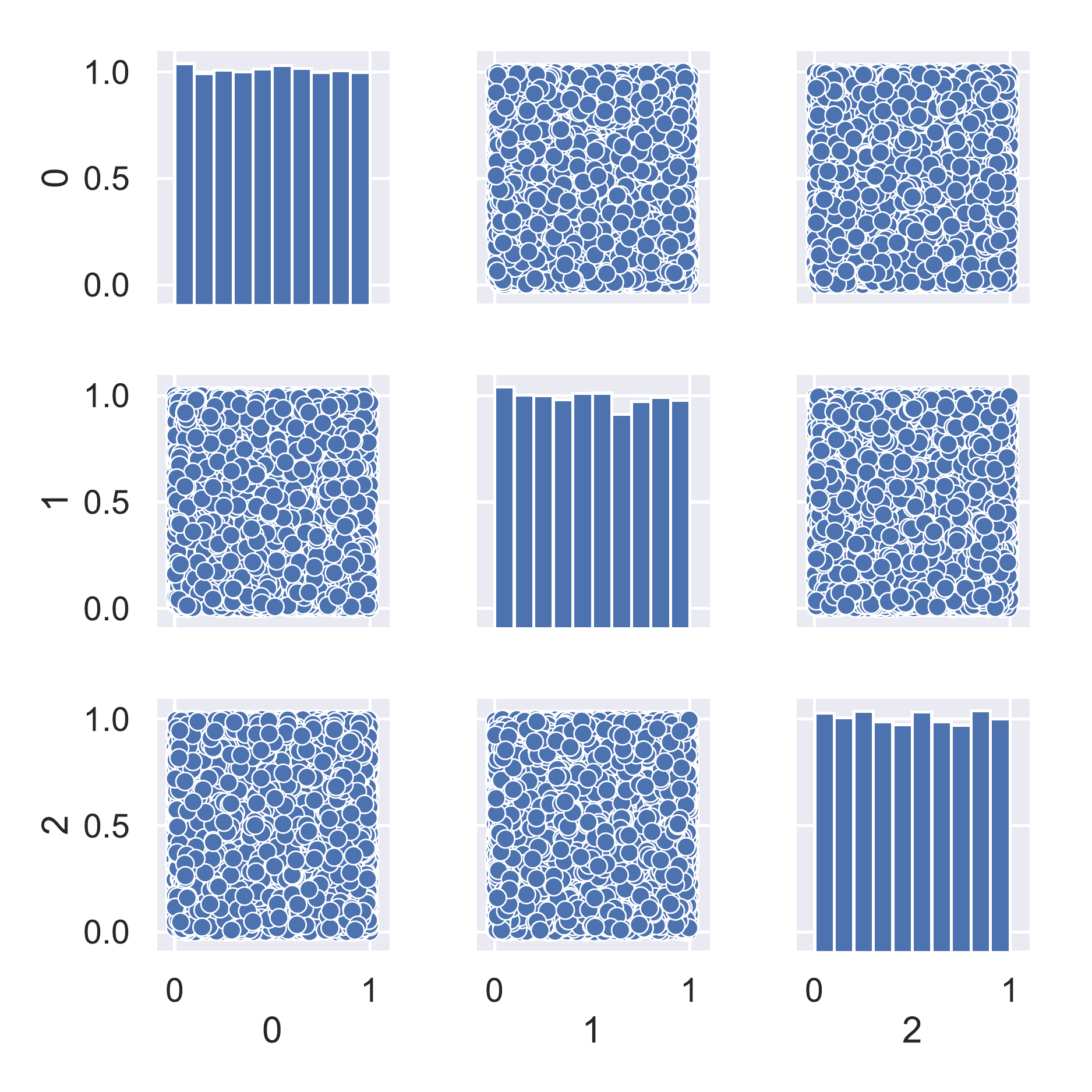}%
		\label{fig:copula_inverse}%
	}
		\caption{The figure depicts inverse flow transformation of the 3
		dimensional toy example in the Section~\ref{sec:experiemnts}. In the
		Figure~\ref{fig:marginal_inverse} we can clearly see the copula
		structures in the off-diagonal scatter plots. The almost uniform
		distribution along the diagonal indicates that $\flow_{\vec X}\inv \approx \vec F_{\vec X} $. The Figure~\ref{fig:copula_inverse} on the right shows inverse transform applied to the output of inverse flow, i.e. $\copulaflow_{\vec X}\inv \left( \flow_{\vec X}\inv \approx \vec F_{\vec C
		X} \right) $. As the evident from scatter plots the copula flow and marginal flow together closely match true CDF see Appendix~\ref{sec:uni_density_approx} }
	\label{fig:inverse_flow}
\end{figure*}    
\begin{figure*}[ht]
	\centering
	\subfloat[Gumbel Copula]{%
		\includegraphics[width=0.33\linewidth]{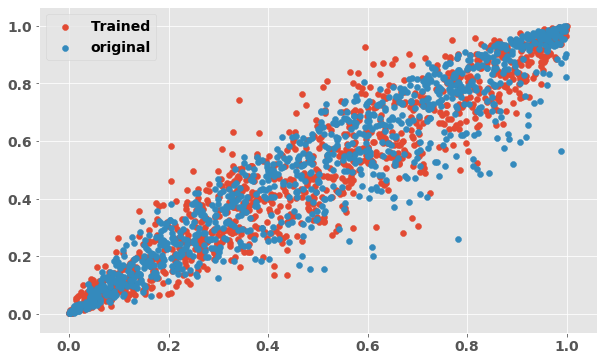}%
		\label{fig:left}%
	} 
	\subfloat[Clayton Copula]{%
		\includegraphics[width=0.33\linewidth]{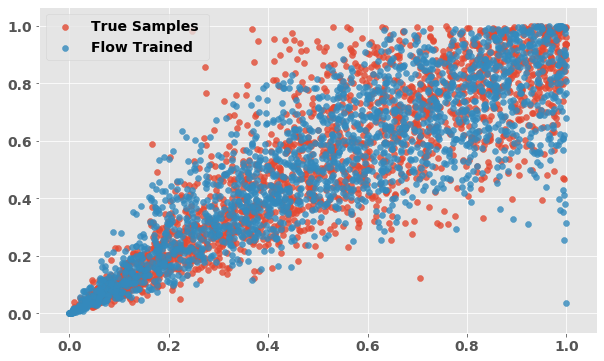}%
		\label{fig:middle}%
	}
	\subfloat[Frank Copula]{%
		\includegraphics[width=0.33\linewidth]{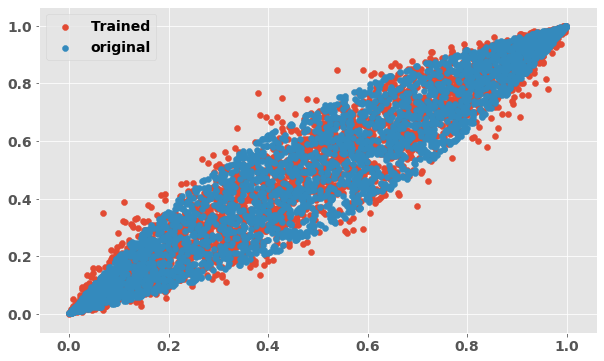}%
		\label{fig:right}%
	}
	\caption{Learning bivariate Copulas using Copula Flow model. 
	Each bi-variate density is learned by the same flow }
	\label{fig:default}
\end{figure*}    

\section{Definitions}

\begin{defn}
	\label{defn:CDF}
A distribution function (cumulative distribution function) is a function $F$
with domain $R = [-\infty, \infty]$, such that 
\begin{itemize}
    \item $F$ is non-decreasing 
    \item $F(-\infty)$ = 0 and $F(\infty) = 1$
    \item $F(x) = \prob[X \leq x] \quad \forall x \in  [-\infty, \infty]$
    \item $F$ is right continuous 
\end{itemize}
\end{defn}

\section{Universal Density Approximator}
\subsection{Copula Flows are Universal Density Approximators }
\label{sec:uni_density_approx}

Any invertible non-linear function that can take i.i.d. $ \uniform{1} $ data and map it onto a
density via monotone functions is a universal
approximator~\citep{Hyvarinen1999}.

The copula flow model can be used to represent any distribution function, i.e., 
the model is a universal density approximator. The inverse of the flow model can be
interpreted as a CDF. By the probability integral transform and Theorem 1
in~\citep{Hyvarinen1999}, we know that if $H_X$ is the CDF of the random variable
then $H_X(X)$ is uniformly distributed in the hyper-cube $[0,1]^d$, where
$d $ is the dimension of the vector. 

In the following, we show
that with the combination of marginal and copula flows we can learn any CDF with desired
accuracy. Then the inverse of this CDF, i.e., the quantile function, can be used to
generate random variables distributed according to $X$.

\paragraph{Convergence of the Marginal Flow} 
\begin{prop}
	\label{prop:spline_universality}
	Monotonic rational quadratic splines can universally approximate any
	monotonic function with appropriate derivative approximations, see
	Theorem~3.1 in~\citep{Gregory1982} 
\end{prop}
As discussed in section~\ref{sec:background}, a (quasi) inverse function $F\inv$ can be used to
transform random variable $U \sim \uniform{1}$ to the random variable $X\sim F$. The inverse function is a monotonic
function, and from Proposition~\ref{prop:spline_universality} we can
say that
\begin{lemma}
	\label{lem:spline_CDF}
	Given a flow map $\flow_X: U \rightarrow \tilde{X}$, such that $\flow_X$
	converges point-wise to the true inverse function $F_X\inv$, the transformed
	random variable $\tilde{X} = \flow_X(U)$ converges in distribution to the
	distribution $X = F_X\inv(U)$.
\end{lemma}
Proof: This is Portmanteau's Theorem applied to flow functions based on Proposition~\ref{prop:spline_universality}

\paragraph{Convergence of the Copula Flow}

For any arbitrary ordering of the variables we
have $\vec U = \vec C_{\vec X} (\vec U_{\vec X}) $. We can write 
\begin{align}
	U_{1} & \doteq C_{X_1}(U_{X_1}, \emptyset ) = \prob(U_{X_1} \le u_{X_1} | \emptyset) \approx \copulaflow_{X_1}\inv(U_{X_1}, \emptyset ) \\
	U_{k} & \doteq  C_{X_k}(U_{X_k}, U_{X_{1:k-1}} )\\
			& = \prob(U_{X_k}\le u_{X_k} | U_{X_{1:k-1}} ) \\
			& \approx \copulaflow_{X_k}\inv(U_{X_k}, U_{X_{1:k-1}} ),  \qquad  2 < k < d, 
\end{align} where $\emptyset$ indicates null set.
According to~\citep{Hyvarinen1999}, the variables $U_1, \ldots, U_d$ are
independently and uniformly distributed. The conditional copula flow is also
modelled as a rational quadratic spline. From Lemma~\ref{lem:spline_CDF}
the $ \copulaflow_{{\vec X}}\inv(\vec U_{\vec X} )$ converges to the true
copula CDF $\vec C_{\vec X}$. Therefore, $\vec \copulaflow_{\vec X} (\vec U)$
converges in distribution to $\vec U_X$.

We can extend the conditional copula flow by adding the marginal flow $U_{X_k}=
F_{X_k}(X_k)$ to obtain 
\begin{align}
	U_{k} & \doteq  C_{X_k}(F_{X_k}(X_k), F_{X_{1:k-1}}(X_{1:k-1}) ) = H_{X_k}(X_k, X_{1:k-1} )\\
			& = \prob(X_k \le x_k | X_{1:k-1} ) \\
			& \approx \copulaflow_{X_k}\inv(\flow_{X_k}\inv (X_k), \flow_{X_{1:k-1}}\inv (X_{1:k-1})),  \qquad  2 < k < d 
\end{align}
From Theorem 1 in~\citep{Hyvarinen1999}, the variables $U_1, \ldots, U_d$ are
independently and uniformly distributed. Moreover, the combined flow is
monotonic as both the copula flow and marginal flow are monotonic. Hence, the
inverse of the combined flow can transform any random vector $\vec X$ to uniform independent
samples in the cube $\uniform{1}^d$.

The flows are learned as invertible monotonic functions. Therefore, using Lemma~\ref{lem:spline_CDF} we obtain that the distribution
$\flow_{\vec X }\inv \left(
\copulaflow_{\vec X} (\vec U) \right)$ converges in distribution to $\vec X$. 
We can generate any random
vector $\vec X$ starting from i.i.d. random variables. Hence, the copula flow model
is universal density approximator.





 \section{Software Details}
 \label{sec:SoftwareDetails}
 A key point of our approach is the interpretation of normalising flow
 functions as a quantile function transforming a uniform density to the desired
 density via inverse transform sampling. As quantile functions are monotonic, we
 use  monotone rational quadratic splines described in~\citep{Durkan2019} as a
 normalising flow function. However, we make changes to the original splines.
 These changes are necessary to ensure that the flow is learning quantile
 function.
 \begin{itemize}
    \item The univariate flow $
\flow  $ maps from the $ \uniform{1} $ to $ \range {F_X} $ of the random variable
$ X $. Hence, unlike~\citep{Durkan2019}, our splines are asymmetric in their
support.
	\item 	We modify the standard spline network to build a map as $ (0, 1)
	\rightarrow (B_{\text{lower}}, B_{\text{upper}}) $ where $B$ is the range of
	the marginal. Infinite
	support can be added with $B\to\infty$.
	\item  We do not parametrise the knot positions by a neural network. We instead
	treat them as a free vector, i.e.,  $ \vec \theta_d  = [\vec \theta_d^w, \vec
	\theta_d^h, \vec \theta_d^s]$ are the width, height and slope parameters,
	respectively, that characterise the flow
	function for the independent marginals of data dimension $\flow_{\vec \theta_d} $
	\item  We map out-of-bound values back into the range and set the gradients
	of the flow map to $0$ at these locations.
\end{itemize}

For copula flow we use same network as autoregressive neural spline
flows~\citep{Durkan2019}. As copula is a CDF defined over uniform densities, the
copula flow spline maps $\uniform{1} \rightarrow \uniform{1}$. Apart from the
change in the range of the splines, the copula flow architecture is sames as
that of autoregressive neural spline flows~\citep{Durkan2019}. 

\begin{table}[]
	\scalebox{0.90}{
	\begin{tabular}{llccccc}
	\hline
				  &                 & Power & Gas   & Hepmass & Miniboone & Synthetic Data \\ \hline
				  & Dimension       & 6     & 8     & 21      & 43        & Various        \\
	 & Train Data Points & 1,615,917      & 852,174        & 315,123        & 29,556         & Various        \\ \hline
				  & Batch Size      & 2048  & 1024  & 1024    & 512       & 1024           \\
	Marginal Flow & Epochs          & 50    & 50    & 50      & 50        & 100            \\
				  & Bins            & 512   & 512   & 512     & 512       & 512            \\
				  & Learning Rate   & 1e-3  & 1e-3  & 1e-3    & 1e-3      & 1e-3           \\ \hline
				  & Batch Size      & 512   & 512   & 512     & 512       & 512            \\
	Copula Flow   & Epochs          & 100   & 100   & 100     & 50        & 100            \\
				  & Bins            & 16    & 16    & 16      & 8         & 16             \\
				  & Learning Rate   & 1e-4  & 1e-4  & 1e-4    & 1e-4      & 1e-4           \\
	 & Hidden Features   & [512, 512] & [256, 256]  & [256, 256] & [128, 128] & [512, 512] \\
				  & Number of Flows & 5     & 10    & 15      & 10        & 10             \\ \hline
	\end{tabular}
	}
	\caption{Hyperprameters for the density estimation results }
	\label{tab:hyper-parameters}
	\end{table}

\end{document}